\documentclass[preprint, 12pt]{elsarticle}
\usepackage{amsmath}

\usepackage{amssymb}
\usepackage{amsthm}
\usepackage{bm}
\let\oldproofname=\proofname
\renewcommand{\proofname}{\rm\bf{\oldproofname}}
\usepackage{tikz}
\usetikzlibrary{shadows,arrows}
\pgfdeclarelayer{background}
\pgfdeclarelayer{foreground}
\pgfsetlayers{background,main,foreground}
\usepackage{pgfplots}
\pgfplotsset{compat=1.3}
\usepackage{hyperref}
\usepackage{graphicx}
\usepackage{pdfpages}
\usepackage{subcaption}
\usepackage{algorithmic}
\usepackage{algorithm}
\usepackage{multirow}
\usepackage{makecell}
\usepackage[font=small,skip=0pt]{caption}
\usepackage{float}
\usepackage{booktabs}
 \makeatletter
 \def\ps@pprintTitle{%
 	\let\@oddhead\@empty
 	\let\@evenhead\@empty
 	\let\@oddfoot\@empty
 	\let\@evenfoot\@oddfoot
 }
 \makeatother
\usepackage{siunitx}
\usepackage{ wasysym }
\usepackage{ textcomp }

\newdefinition{rmk}{Remark}
\newproof{pf}{Proof}
\newdefinition{defi}{Definition}
\usepackage{tablefootnote}
\begin{document}	
\begin{frontmatter}

\title{Unsupervised Feature Selection based on Adaptive Similarity Learning and Subspace Clustering}

\author[rv1]{Mohsen Ghassemi Parsa}
\author[rv1]{Hadi Zare \corref{cor1}}
\author[rv2]{Mehdi Ghatee}
\cortext[cor1]{Corresponding author}
\address[rv1]{Faculty of New Sciences and Technologies, University of Tehran, Iran}
\address[rv2]{Department of Computer Science, Amirkabir University of Technology, Iran}
\begin{abstract}
Feature selection methods have an important role on the readability of data and the reduction of complexity of learning algorithms. In recent years, a variety of efforts are investigated on feature selection problems based on unsupervised viewpoint due to the laborious labeling task on large datasets. 
In this paper, we propose a novel approach on unsupervised feature selection initiated from the subspace clustering to preserve the similarities by representation learning of low dimensional subspaces among the samples. A self-expressive model is employed to implicitly learn the cluster similarities in an adaptive manner.
The proposed method not only maintains the sample similarities through subspace clustering, but it also captures the discriminative information based on a regularized regression model. In line with the convergence analysis of the proposed method, the experimental results on benchmark datasets demonstrate the effectiveness of our approach as compared with the state of the art methods.
\end{abstract}

\end{frontmatter}

\section{Introduction}
\label{intro}
One of the most common approaches for dealing with high dimensional data is to select the appropriate features which is known as feature selection (FS) problem in machine learning community \cite{guyon_introduction_2003, li_feature_2017}. FS techniques are widely applied in many domains including text mining \cite{liu_evaluation_2003}, bioinformatics \cite{ding_chris_h.q._unsupervised_2003}, social media \cite{tang_unsupervised_2014}, and ensemble learning \cite{abpeykar_ensemble_2019}. On the one hand, FS approach provides a sparse representation for data with massive number of features to alleviate the curse of dimensionality effect on the learning performance \cite{bishop_pattern_2006}. On the other hand, the computational burden of facing with massive data could be decreased significantly through FS approach as compared with other techniques like feature extraction approaches \cite{guyon_feature_2006}.

FS techniques can be categorized into wrapper, filter and embedded approaches by considering the evaluation criteria. In wrapper methods \cite{dy_feature_2004, kohavi_wrappers_1997, roth_feature_2003}, the evaluation process depends on the learning algorithms, in contrast to the filter methods \cite{mitra_unsupervised_2002, he_laplacian_2005, zhao_spectral_2007} that only use data for evaluating without any learning phase. The embedded methods \cite{hou_joint_2014, han_unsupervised_2020}, embed the process of selecting features in a learning algorithm.

On learning taxonomy, FS problem can be stated as ``Supervised'' and ``Unsupervised''. Supervised FSs are primarily constructed from the dependency among the features and the label information, including information theoretic approaches \cite{peng_feature_2005}, statistical tests \cite{hall_feature_1999, huan_liu_chi2:_1995}, sparse learning methods \cite{liu_multi-task_2009,nie_efficient_2010}, and structure learning \cite{zare_relevant_2016}.
 On the other hand,  appropriate criterion on Unsupervised FS (UFS) problems is more challenging due to the ill-defined nature of the problem. There have been a variety of works on UFS \cite{solorio-fernandez_review_2019} including metaheuristic approach \cite{tabakhi_unsupervised_2014}, graph clustering \cite{moradi_graph_2015}, feature-level reconstruction \cite{li_reconstruction-based_2017}, discriminative approach \cite{yang_l21-norm_2011}, and spectral clustering \cite{li_unsupervised_2012, li_clustering-guided_2014}.
One of the important aims in UFS approaches is to preserve the geometric structure in the selected features \cite{he_laplacian_2005}. 
To this end, the similarity preserving methods construct a graph similarities to maintain the geometric structure in the reduced space \cite{liu_global_2014}.

The graph similarity computation is often performed independently from the feature selection, which may lead to a suboptimal solution. Adaptive structure learning methods \cite{du_unsupervised_2015} explicitly learn a similarity matrix which expand the search space of solutions. 
There are some UFS by considering the subspace learning to exploit the hidden multidimensional substructures of data \cite{wang_subspace_2015}. The main idea of the subspace learning is constructed from the representation of high dimensional data points based on the union of the subspaces \cite{soltanolkotabi_robust_2014}.
Subspace clustering refers to the process of clustering and detecting the low dimensional structure of the clusters at the same time \cite{vidal_subspace_2011}.
 The earlier subspace learning methods \cite{shang_subspace_2016, shang_local_2019} have considered the self-expressiveness of the features aligned with the graph similarity stage. 
 
 \begin{figure}[t]
 	\centering
 	\includegraphics[width=0.9\textwidth]{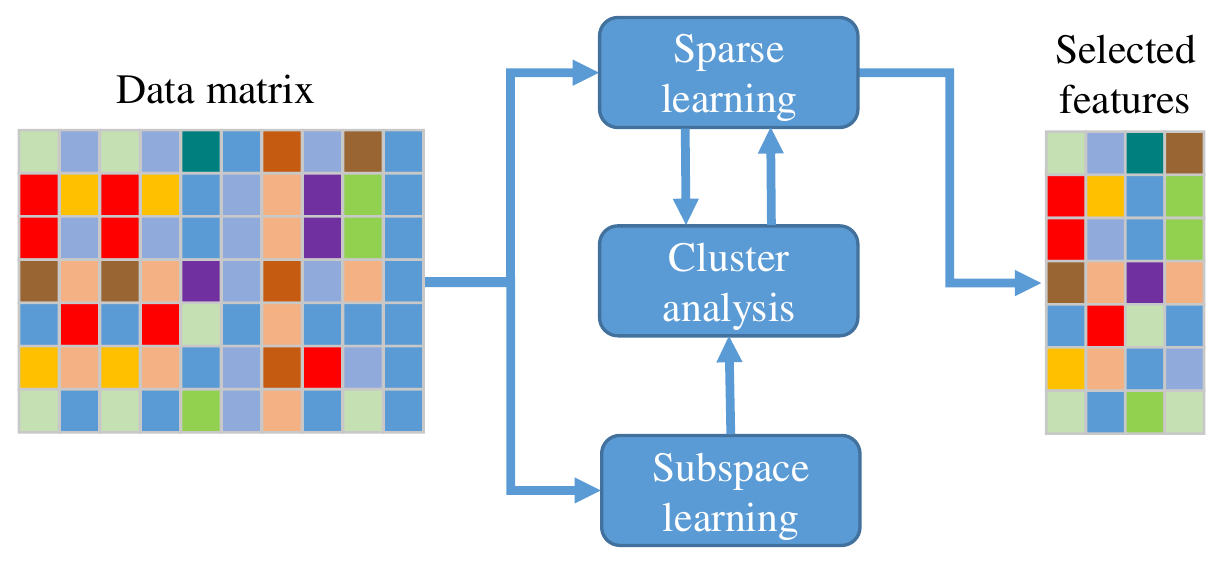}
 	\caption{The outline of the proposed method.}
 	\label{fig:Fig1}
 \end{figure} 

In this paper, we propose a novel UFS, ``Subspace Clustering unsupervised Feature Selection'' (SCFS), to exploit the discriminative information concurrent with cluster analysis and adaptively maintain the similarity structure through an implicit similarity matrix computation.
The proposed approach is constituted by a self-expressive model to include both of learning latent homogeneous structures and similarity matrix computation in a unified objective function aligned with an $\ell_{2,1}$-norm to address a regularized regression model. 
In our approach, subspace learning is utilized to maintain the cluster similarities in the selected features and sparse learning is applied to learn a regularized regression model to measure the correlation between the features and the learned clustering information. The more a feature is related to the clusters, the more it is likely to be selected. The outline of the proposed method is presented in \mbox{Fig. \ref{fig:Fig1}}. 

The main contributions of this paper are as follows,
\begin{itemize}
	\item We propose a novel UFS method by applying subspace learning, cluster analysis and sparse learning to consider the sample similarities and discriminative information in the selected features.
	\item We introduce a self-expressive model to adaptively and implicitly learn the cluster similarities.
	\item We use a regularized regression approach to compute the sparse correlation among the features and clusters.
	\item We introduce an optimization algorithm to address the proposed objective function.
\end{itemize}
This paper is organized as follows. The related works are introduced in \mbox{Section \ref{sec:lit_rew}}. 
The proposed method and the corresponding optimization algorithm are presented in
\mbox{Section \ref{sec:propos}}. Convergence analysis and computational complexity are discussed in \mbox{Section \ref{sec:discus}}. Experimental setting and results are reported in \mbox{Section \ref{sec:exp}}. Finally, the conclusion of the paper is provided in \mbox{Section \ref{sec:concl}}.

\section{Related Works}\label{sec:lit_rew}

Unsupervised feature selection methods generally select features based on the intrinsic structural characteristics of data. These methods can be divided into three main categories, similarity preserving \cite{he_laplacian_2005}, data reconstruction \cite{masaeli_convex_2010}, and sparse learning approaches \cite{li_unsupervised_2012}.

The main focus of similarity preserving methods including Laplacian Score \cite{he_laplacian_2005} and SPEC \cite{zhao_spectral_2007}, is to maintain the local similarities. Reconstruction based methods employ a feature-level self-expressive model including convex principal feature selection,  CPFS \cite{masaeli_convex_2010}, embedded reconstruction based, REFS \cite{li_reconstruction-based_2017} and structure preserving, SPUFS \cite{lu_structure_2018}. 
Sparse unsupervised FS approaches are initiated from the ideas of the sparse machine learning \cite{bach_optimization_2012}. The core idea is to embed FS in a regularized learning model. There are several well-known approaches in this category including preserving the multi-cluster structure of data, MCFS \cite{cai_unsupervised_2010}, local discriminative approach using the scatter matrix, UDFS \cite{yang_l21-norm_2011}, joint embedding learning and sparse regression, JELSR \cite{hou_joint_2014}, nonnegative discriminative feature selection based on spectral clustering, NDFS \cite{li_unsupervised_2012}, global similarity preserving feature selection, SPFS \cite{zhao_similarity_2013}, global and local similarity preserving feature selection, GLSPFS \cite{liu_global_2014}, and unsupervised feature selection with adaptive structure learning, FSASL \cite{du_unsupervised_2015}.
 
There are some UFS methods by considering the subspace learning idea. Feature-level reconstruction based approach was proposed in, MFFS \cite{wang_subspace_2015} by exploiting the matrix factorization.  
In \cite{shang_subspace_2016}, a graph regularized approach was introduced to maintain the local similarities. A sparse discriminative learning approach was devised in \cite{shang_local_2019} to select discriminative features based on the local structure of the samples. While, UFS methods with the aid of subspace learning, MFFS \cite{wang_subspace_2015}, SGFS \cite{shang_subspace_2016}, LDSSL \cite{shang_local_2019} mainly reconstruct the data matrix in feature-level, the sample-level characteristics such as the cluster structures are not thoroughly incorporated in them. 
%%%%%%%%%%%%%%%%%%%%%%%%%%%%%
\begin{table}[t]
	\centering
	\tiny
	\caption{A comparison of the related unsupervised feature selection methods.}
	%{\renewcommand{\arraystretch}{1.2}
	\begin{tabular}{l c c c c c c c}
		\toprule%[1.5pt]
		{Algorithm} & \makecell{Self-\\expression} & \makecell{Similarity preserving} & \makecell{Adaptive\\graph matrix} &  \makecell{Joint\\learning} & \makecell{Cluster\\analysis}  & \makecell{Regression} & \makecell{Regularization}\\ 
		\midrule
		LS \cite{he_laplacian_2005}				&\texttimes & Explicit	&\texttimes  &\checkmark &\texttimes &\texttimes &\texttimes\\
		CPFS \cite{masaeli_convex_2010} &Feature &\texttimes	&\texttimes  &\texttimes &\texttimes &\texttimes &\checkmark\\
		MCFS \cite{cai_unsupervised_2010}  &\texttimes &Explicit	 &\texttimes  &\texttimes &\texttimes	 &\checkmark &\checkmark\\ 
		UDFS \cite{yang_l21-norm_2011}		&\texttimes &Explicit	 &\texttimes  &\checkmark &\texttimes &\checkmark &\checkmark\\ 
		NDFS \cite{li_unsupervised_2012}  &\texttimes &Explicit	&\texttimes  &\checkmark &\checkmark&\checkmark&\checkmark\\ 
		GLSPFS \cite{liu_global_2014}	 &\texttimes	&Explicit	&\texttimes  &\checkmark &\texttimes&\checkmark&\checkmark\\
		MFFS \cite{wang_subspace_2015} &Feature  &\texttimes	&\texttimes &\texttimes &\texttimes &\texttimes &\texttimes\\
		FSASL \cite{du_unsupervised_2015}  &\texttimes &Explicit	&\checkmark &\checkmark &\texttimes&\checkmark&\checkmark\\
		REFS \cite{li_reconstruction-based_2017}	&Feature	&Explicit	&\texttimes	 &\checkmark &\texttimes &\texttimes&\texttimes\\
		SPUFS \cite{lu_structure_2018}	&Feature	&Explicit	&\texttimes	 &\checkmark &\texttimes &\texttimes&\checkmark\\
		LDSSL \cite{shang_local_2019}	&Feature	&Explicit	&\texttimes  &\checkmark &\texttimes &\texttimes&\checkmark\\
		SCFS 										&Sample	&Implicit	&\checkmark	&\checkmark&\checkmark&\checkmark&\checkmark\\ 
		\bottomrule%[1.5pt] 
	\end{tabular}%}
	\label{tb_comp}	
\end{table}

The important properties of the well-known UFS methods are summarized in \mbox{Table \ref{tb_comp}}. 
Theses methods are compared based on multiple properties including, Self-expression, Similarity preserving, Adaptive graph matrix, Joint learning, Cluster analysis, Regression and Regularization. Self-expressive property points out the reconstruction on samples or features. Similarity preserving is related to maintain sample similarities and computing explicitly or implicitly of the similarity matrix. Adaptive graph matrix indicates the property of learning the similarity matrix 
 concurrent with the feature selection process. Joint learning refers to perform both of subspace learning and feature selection in a unified framework. Cluster analysis indicates that a method employs any clustering algorithm to select relevant features. Regression refers to exploit a regression model to discover discriminative features. Finally, Regularization indicates the consideration of regularization factors in the method to result a sparse solution. Most of the UFS techniques are constructed from one or more of these properties, but in this work, a unified framework is proposed to consider all of the characteristics to provide a more robust UFS.

\section{The Proposed Method}\label{sec:propos}
\subsection{Notations}
Throughout this paper, matrices are denoted by bold uppercase and vectors by bold lowercase characters. Let $\mathbf{B}$ be an arbitrary matrix , $B_{ij}$ is its ${(i,j)}$-th element, and $\mathbf{b_{i}}$ denotes the $i$-th row. The Frobenius norm, the trace and the transpose operators on matrix $\mathbf{B}$ are denoted by ${\lVert \mathbf{B} \rVert}_F$, $\textrm{tr}(\mathbf{B})$, and $\mathbf{B}^{\top}$, respectively. The $\ell_{2}$-norm of a vector $\mathbf{v}$ is denoted as ${\lVert \mathbf{v} \rVert}_2$ and the $\ell_{2,1}$-norm is defined as following,
\begin{equation*}
{\lVert \mathbf{B} \rVert}_{2,1}= \sum\limits_{i}\sqrt{\sum\limits_{j}B_{ij}^2}.
\end{equation*}

$\mathbf{X}\in \mathbb{R}^{n\times p}$ denotes the data matrix, where $n$ and $p$ are the number of samples and features. $\mathbf{G}\in \mathbb{R}^{n\times c}$ represents the clustering matrix, where $c$ is the number of clusters. 

\subsection{The Proposed Method}
At first, the similarity matrix is implicitly computed by subspace learning. The proposed self-expressive similarity representation is given in \mbox{Eq. \eqref{eq:3}},
\begin{equation}
\label{eq:3}
\begin{array}{r l} 
\displaystyle \min_{\mathbf{G}} &{\lVert \mathbf{X} - \mathbf{G}\mathbf{G}^{\top}\mathbf{X} \rVert}_F^2 \\[10pt]
\textrm{s.t.} & \mathbf{G}\geq 0,\mathbf{G}\mathbf{G}^{\top}\mathbf{1} = \mathbf{1},
\end{array}
\end{equation}
where $\mathbf{1}$ is an $n \times n$ matrix of ones and the constraint $\mathbf{G}\mathbf{G}^{\top}\mathbf{1} = \mathbf{1}$ is imposed to normalize the similarity matrix.
The symmetric nonnegative matrix $\mathbf{G}\mathbf{G}^{\top}$ is learned such that the samples within common subspaces tend to attain large values in $\mathbf{G}\mathbf{G}^{\top}$. In lines with a low dimensional representation of $\mathbf{G}$ by assuming $c < \{n, p\}$, $\mathbf{G}$ can also be interpreted as a clustering matrix.
Moreover, $\mathbf{G}\mathbf{G}^{\top}$, represents the pairwise sample similarities in terms of the clustering values. 
\par The next stage is to construct a sparse transformation $\mathbf{W}$ on the data matrix $\mathbf{X}$ by employing the clustering matrix $\mathbf{G}$, joined with a regularization term,
\begin{equation}\label{eq:7}
\begin{array}{r l} 
\displaystyle \min_{ \mathbf{W}} & {\lVert \mathbf{XW} - \mathbf{G} \rVert}_F^2 + \beta {\lVert \mathbf{W} \rVert}_{2,1},
\end{array}
\end{equation}
where $\mathbf{W}\in \mathbb{R}^{p\times c}$ is a linear and low dimensional transformation matrix, and $\beta$ is a regularization
 parameter.
The objective function in \mbox{Eq. \eqref{eq:7}} represents the linear transformation model to measure the association between features and clusters.
The $\ell_{2,1}$ norm induces sparsity on the rows of the transformation matrix, $\mathbf{w_i}$'s. When $\mathbf{w_i}$'s are closer to zero, their correspondence features are less relevant and more likely to be eliminated from the final candidate set of the discriminative features.

By integrating \mbox{Eq. \eqref{eq:3} and \eqref{eq:7}} in a joint objective function, our final model is obtained as follows,
\begin{equation}
\label{eq:8}
\begin{array}{r l} 
\displaystyle \min_{\mathbf{W},\mathbf{G}} &{\lVert \mathbf{X} - \mathbf{G}\mathbf{G}^{\top}\mathbf{X} \rVert}_F^2 + \alpha {\lVert \mathbf{XW} - \mathbf{G} \rVert}_F^2 + \beta {\lVert \mathbf{W} \rVert}_{2,1}
\\[10pt]
\textrm{s.t.} & \mathbf{G}\geq 0,\mathbf{G}\mathbf{G}^{\top}\mathbf{1} = \mathbf{1},
\end{array}
\end{equation}
where $\alpha$ is a tuning parameter. 
\begin{figure}[t]
	\centering
	\includegraphics[width=\textwidth]{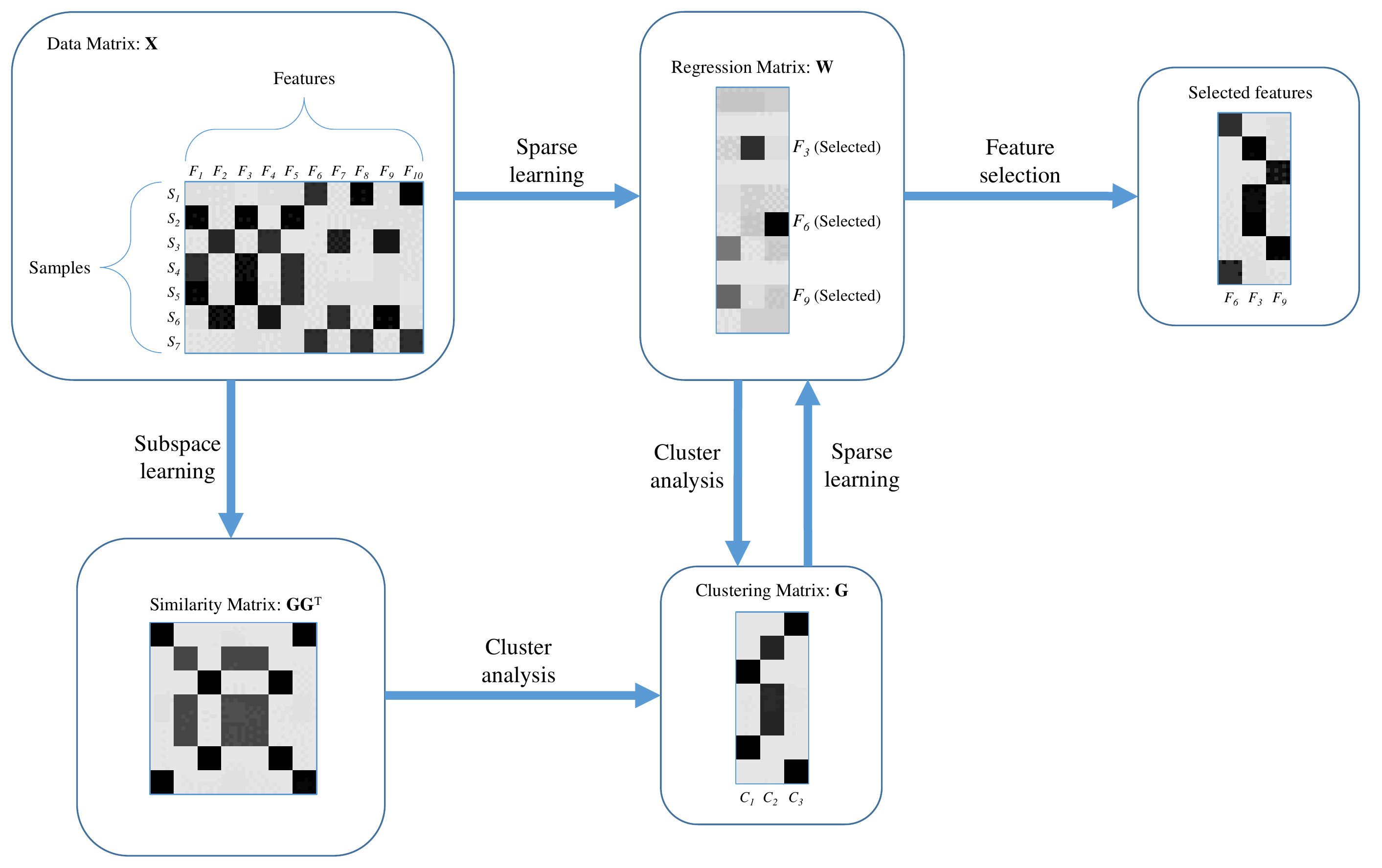}
	\caption{The description of the proposed method.}
	\label{fig:Fig2}
\end{figure}
By solving the objective function in \mbox{Eq. \eqref{eq:8}}, $\mathbf{W}$ and $\mathbf{G}$ are iteratively updated in advance of achieving the optimal result.

 We illustrate the steps of SCFS in \mbox{Fig. \ref{fig:Fig2}}. $\mathbf{X}$ is a nonnegative artificial data matrix with seven samples and ten features. The bright entries of $\mathbf{X}$ are close to zero and the dark ones are far from zero. Initially, subspace learning stage provides the cluster similarities $\mathbf{G}$ which is used to construct the similarity matrix $\mathbf{G}\mathbf{G}^{\top}$ among samples.
Then, a sparse learning method is applied for learning the regularized coefficients $\mathbf{W}$ through a regression model to measure the importance of features. The $\mathbf{W}$ and $\mathbf{G}$ are optimized in an iterative process. Finally, the most important features are selected based on $\mathbf{W}$.  
In this example, $F_6$, $F_3$, and $F_9$ are selected according to their roles' in the learned hidden subspaces.
 \subsection{Optimization}
The primary objective function in \mbox{Eq. \eqref{eq:8}} can be considered as,
\begin{equation}\label{eq:11}
\begin{array}{r l} 
 \min\limits_{\mathbf{W},\mathbf{G}\geq 0,\mathbf{G}\mathbf{G}^{\top}\mathbf{1} = \mathbf{1}}&f(\mathbf{W},\mathbf{G})={\lVert \mathbf{X} - \mathbf{G}\mathbf{G}^{\top}\mathbf{X} \rVert}_F^2 + \alpha {\lVert \mathbf{XW} - \mathbf{G} \rVert}_F^2 + \beta {\lVert \mathbf{W} \rVert}_{2,1}.
\end{array}
\end{equation}
A gradient based procedure is utilized to solve this optimization problem by considering the main elements $\mathbf{W}$ and $\mathbf{G}$. It begins by fixing one element and finding the optimum value for the other ones which is described in below.
Initially, $\mathbf{G}$ is fixed to yield the following objective function, 
\begin{equation}\label{eq:9}
\begin{array}{r l} 
 \displaystyle \min_{\mathbf{W}}&f(\mathbf{W})=\alpha {\lVert \mathbf{XW} - \mathbf{G} \rVert}_F^2 + \beta {\lVert \mathbf{W} \rVert}_{2,1}.
\end{array}
\end{equation}
Taking the derivative to calculate the $\nabla f(\mathbf{W})$  and setting it to zero,
\begin{equation}\label{eq:10}
\mathbf{W} = {\left(\alpha\mathbf{X}^{\top} \mathbf{X} + \beta \mathbf{D} \right)}^{-1}\alpha \mathbf{X}^{\top} \mathbf{G},
\end{equation}
where $\mathbf{D}$ is a diagonal matrix with,
\begin{equation}
\label{eq:17}
D_{ii}=\frac{1}{2\lVert \mathbf{w_{i}} \rVert_{2}+\epsilon},
\end{equation}
where $\epsilon$ is a very small positive number to prevent the division by zero.\\
Then, $\mathbf{G}$ is updated through the objective function in Eq. \eqref{eq:12} by fixing  $\mathbf{W}$, 
\begin{equation}\label{eq:12}
\begin{array}{r l} 
\min\limits_{\mathbf{G}\geq 0,\mathbf{G}\mathbf{G}^{\top}\mathbf{1} = \mathbf{1}}&f(\mathbf{G})={\lVert \mathbf{X} - \mathbf{G}\mathbf{G}^{\top}\mathbf{X} \rVert}_F^2 + \alpha {\lVert \mathbf{XW} - \mathbf{G} \rVert}_F^2.
\end{array}
\end{equation}
Eq. \eqref{eq:12} is rewritten to relax the constraints as,
\begin{equation}\label{eq:14}
\begin{array}{c} 
f(\mathbf{G})={\lVert \mathbf{X} - \mathbf{G}\mathbf{G}^{\top}\mathbf{X} \rVert}_F^2 + \alpha {\lVert \mathbf{XW} - \mathbf{G} \rVert}_F^2 + \gamma{\lVert \mathbf{G}\mathbf{G}^{\top}\mathbf{1} - \mathbf{1} \rVert}_F^2 + tr\left(\mathbf{\Phi} \mathbf{G}^{\top}\right),
\end{array}
\end{equation}
where $\gamma>0$ is a parameter to control the normalizing constraint and practically should be a large number. $\mathbf{\Phi}$ is the Lagrange multiplier for $\mathbf{G}\geq 0$ constraint.
Setting the derivative of $f(\mathbf{G})$ with respect to $\mathbf{G}$ to $0$,
\begin{equation}\label{eq:15}
2\mathbf{M}\mathbf{G}^{\top}\mathbf{G}+2\mathbf{G}\mathbf{G}^{\top}\mathbf{M}+2\alpha \mathbf{G}-4\mathbf{M}-2\alpha \mathbf{X}\mathbf{W}+ \mathbf{\Phi} = 0,
\end{equation}
where $\mathbf{M} = \left(\mathbf{X}\mathbf{X}^{\top}+n\gamma\mathbf{1}\right)\mathbf{G}$. By applying the KKT condition \cite{kuhn_nonlinear_2014}, the following updating rule is obtained,
\begin{equation}\label{eq:16}
\begin{array}{c} 
G_{ij} = G_{ij} \cfrac{[2\mathbf{M}+\alpha \mathbf{X}\mathbf{W}]_{ij}}{[\mathbf{M}\mathbf{G}^{\top}\mathbf{G}+\mathbf{G}\mathbf{G}^{\top}\mathbf{M}+\alpha \mathbf{G}]_{ij}},
\end{array}
\end{equation}

Therefore, by initializing the $\mathbf{G}$ and $\mathbf{D}$, 
in each iteration of the proposed formulation, first $\mathbf{W}$ is updated by \mbox{Eq. \eqref{eq:10}}, and then $\mathbf{G}$ and $\mathbf{D}$ is updated by \mbox{Eq. \eqref{eq:16} and \eqref{eq:17}}. \mbox{Algorithm \ref{alg1}} describes the optimization process of the proposed method.
\begin{algorithm}[h!] % enter the algorithm environment
	\floatname{algorithm}{Algorithm}
	\renewcommand{\algorithmicrequire}{\textbf{Input:}}
	\renewcommand{\algorithmicensure}{\textbf{Output:}}
	\caption{SCFS algorithm.} % give the algorithm a caption
	\label{alg1} % and a label for \ref{} commands later in the document
	\begin{algorithmic}[1] % enter the algorithmic environment
		\REQUIRE Data matrix $\mathbf{X}\in \mathbb{R}^{n\times p}$ and parameters $\alpha$ and $\beta$.
		\STATE $t=0$.
		\STATE Initialize $\mathbf{G}_0 \in \mathbb{R}^{n\times c}$.
		\STATE Initialize $\mathbf{D}_0$ as an identity matrix.
		\REPEAT
		\vspace{4pt}
		\STATE $\mathbf{W}_{t+1} = {\left(\alpha\mathbf{X}^{\top} \mathbf{X} + \beta \mathbf{D}_{t} \right)}^{-1}\alpha \mathbf{X}^{\top} \mathbf{G}_{t}$.
		\vspace{4pt}
		\STATE $\mathbf{M}_{t} = \left(\mathbf{X}\mathbf{X}^{\top}+n\gamma\mathbf{1}\right)\mathbf{G}_{t}$. 
		\vspace{4pt}
		\STATE $(G_{t+1})_{ij} = (G_{t})_{ij} \cfrac{[2\mathbf{M}_{t}+\alpha \mathbf{X}\mathbf{W}_{t+1}]_{ij}}{[\mathbf{M}_{t}\mathbf{G}_{t}^{\top}\mathbf{G}_{t}+\mathbf{G}_{t}\mathbf{G}_{t}^{\top}\mathbf{M}_{t}+\alpha \mathbf{G}_{t}]_{ij}}$.
		\vspace{4pt}
		\STATE Update the diagonal matrix $\mathbf{D}$ as 
		$(D_{t+1})_{ii}=\frac{1}{2\lVert (\mathbf{w}_{t+1})_{i} \rVert_{2}+\epsilon}$.
		\vspace{4pt}
		\STATE $t = t + 1$.
		\vspace{4pt}
		\UNTIL{Convergence of the objective function in \mbox{Eq. \eqref{eq:8}}}.
		\ENSURE Sort features by descending order of ${\lVert \mathbf{w_{i}} \rVert}_2$.
	\end{algorithmic}
\end{algorithm}

\section{The analysis of the proposed algorithm}\label{sec:discus}
This section presents the convergence behavior and computational complexity of SCFS.
\subsection{Convergence Analysis}\label{subsec:conv_ana}
Our aim is to show the non-increasing behavior of the primary objective function in \mbox{Eq. \eqref{eq:8}}. First, a lemma is given \cite{nie_efficient_2010}.
\newtheorem{thm2}{Lemma}
\begin{thm2}\label{lem:1}
	For any nonzero vectors $\mathbf{u}, \mathbf{v} \in \mathbb{R}^p$, the following holds,
	\begin{align}	
	\lVert \mathbf{u} \rVert_{2} - \frac{{\lVert \mathbf{u} \rVert}_{2}^2}{2\lVert \mathbf{v} \rVert_{2}}
	\leq
	\lVert \mathbf{v} \rVert_{2} - \frac{{\lVert \mathbf{v} \rVert}_{2}^2}{2\lVert \mathbf{v} \rVert_{2}}.
	\end{align}
\end{thm2}
\newtheorem{thm1}{Theorem}
\begin{thm1}
	The objective function in \mbox{Eq. \eqref{eq:8}} is non-increasing in each iteration by employing the updating rules in \mbox{Algorithm \ref{alg1}}.
\end{thm1}
\begin{proof}
	First, the objective function can be written as,
	\begin{equation}\label{eq:20}
	\begin{array}{r l} 
	f(\mathbf{W},\mathbf{G})=&{\lVert \mathbf{X} - \mathbf{G}\mathbf{G}^{\top}\mathbf{X} \rVert}_F^2 + \alpha {\lVert \mathbf{XW} - \mathbf{G} \rVert}_F^2 +\beta {\lVert \mathbf{W} \rVert}_{2,1}\\
	& +\gamma{\lVert \mathbf{G}\mathbf{G}^{\top}\mathbf{1} - \mathbf{1} \rVert}_F^2.
	\end{array}
	\end{equation}
By fixing $\mathbf{G}_{t}$, we should justify the following inequality, 
	\begin{equation}
	f(\mathbf{W}_{t+1},\mathbf{G}_{t}) \leq f(\mathbf{W}_{t},\mathbf{G}_{t}).
	\label{eq:W}
	\end{equation}
Based on Eq. \eqref{eq:9},  inequality \eqref{eq:W} can be written as,	
	\begin{align}	
	\begin{array}{l}
	{\lVert \mathbf{X}\mathbf{W}_{t+1} - \mathbf{G}_{t} \rVert}_F^2 + \beta \sum_{i=1}^{p}( \frac{{\lVert (\mathbf{w}_{t+1})_{i} \rVert}_2^2}{2{\lVert (\mathbf{w}_{t})_{i} \rVert}_2})\\[10pt]
	\leq 
	{\lVert \mathbf{X}\mathbf{W}_{t} - \mathbf{G}_{t} \rVert}_F^2 + \beta \sum_{i=1}^{p}( \frac{{\lVert (\mathbf{w}_{t})_{i} \rVert}_2^2}{2{\lVert (\mathbf{w}_{t})_{i} \rVert}_2} ).
	\end{array}
	\label{eq:W3}
	\end{align}
The inequality  \eqref{eq:W3} is followed as,
	\begin{align}	
	&{\lVert \mathbf{X}\mathbf{W}_{t+1} - \mathbf{G}_{t} \rVert}_F^2 + \beta {\lVert \mathbf{W}_{t+1} \rVert}_{2,1} - \beta \sum\limits_{i=1}^{p}({\lVert (\mathbf{w}_{t+1})_{i} \rVert}_2 - \frac{{\lVert (\mathbf{w}_{t+1})_{i} \rVert}_2^2}{2{\lVert (\mathbf{w}_{t})_{i} \rVert}_2}	)\nonumber\\
 & \leq
	{\lVert \mathbf{X}\mathbf{W}_{t} - \mathbf{G}_{t} \rVert}_F^2 + \beta {\lVert \mathbf{W}_{t} \rVert}_{2,1} - \beta \sum\limits_{i=1}^{p}({\lVert (\mathbf{w}_{t})_{i} \rVert}_2 - \frac{{\lVert (\mathbf{w}_{t})_{i} \rVert}_2^2}{2{\lVert (\mathbf{w}_{t})_{i} \rVert}_2} ).
	\label{eq:22}
	\end{align}
	According to \mbox{Lemma \ref{lem:1}},
	\begin{equation}	
	{\lVert \mathbf{X}\mathbf{W}_{t+1} - \mathbf{G}_{t} \rVert}_F^2 + \beta {\lVert \mathbf{W}_{t+1} \rVert}_{2,1} 
	\leq 
	{\lVert \mathbf{X}\mathbf{W}_{t} - \mathbf{G}_{t} \rVert}_F^2 + \beta {\lVert \mathbf{W}_{t} \rVert}_{2,1}.
	\label{eq:w2}
	\end{equation}
Taking fixed $\mathbf{W}_{t+1}$, based on a similar approach in \cite{lee_algorithms_2000}, it follows,
\begin{equation}
f(\mathbf{W}_{t+1},\mathbf{G}_{t+1}) \leq f(\mathbf{W}_{t+1},\mathbf{G}_{t}).
\label{eq:G}
\end{equation}	
Hence,
\begin{equation}
f(\mathbf{W}_{t+1},\mathbf{G}_{t+1}) \leq f(\mathbf{W}_{t+1},\mathbf{G}_{t})  \leq f(\mathbf{W}_{t},\mathbf{G}_{t}).
\end{equation}
Therefore, Algorithm \ref{alg1} will monotonically decrease the objective function in Eq. \eqref{eq:8} based on the relations \eqref{eq:w2} and \eqref{eq:G}.
\end{proof}
%Therefore, Algorithm \ref{alg1} will monotonically decrease the objective function in Eq. \eqref{eq:8} based on the relations \eqref{eq:w2} and \eqref{eq:G}.
\subsection{Computational complexity}
The main steps of \mbox{Algorithm \ref{alg1}} contains the updating $\mathbf{W}$ and $\mathbf{G}$ on each iteration.
The update of $\mathbf{W}$ and $\mathbf{G}$ take $O(p^3+np^2+npc)$ and $O(n^2p+n^2c+npc)$ time complexity.
 Hence, the time complexity of the proposed algorithm is $\max\{O(p^3), O(np^2), O(n^2p), O(n^2c), O(npc)\}$. In most applied scenarios $c\ll p$, that implies the time complexity of the proposed algorithm could be reduced to, $\max\{O(p^3) , O(n^2p)\}$. 
\section{Experiments}\label{sec:exp}
In this section, the proposed method is evaluated using benchmark datasets by standard evaluation measures. A bunch of state-of-the-art FS methods are compared with SCFS where the results and experimental setting are reported in the following.
\subsection{Datasets}
A variety of datasets are applied in different domains including biological (Lung, Lymphoma, Prostate-GE), image (ORL), voice (Isolet), and text (BASEHOCK) data. All of the datasets are available on repository \cite{li_feature_2017}. \mbox{Table \ref{tb_datasets}} reports the main characteristics of datasets. 
\begin{table}[t]
	\centering
	\footnotesize
	\caption{The main properties of datasets in the experiments.}
	\begin{tabular}{ l c c c c c}
		\toprule%[1pt]
		Dataset & n & p &c & \multicolumn{1}{c}{Type} & \multicolumn{1}{c}{Domain}\\
		\midrule 
		Lung & 203 & 3312 & 5 & Continuous & Biology \\
		Lymphoma & 96 & 4026 & 9 & Discrete & Biology \\
		Prostate-GE & 102 & 5966 & 2 & Continuous & Biology \\
		ORL & 400 & 1024 & 40 & Discrete & Image \\		
		Isolet & 1560 & 617 & 26 & Continuous & Voice \\
		BASEHOCK & 1993 & 4862 & 2 & Discrete & Text \\
	\end{tabular}\\
	\label{tb_datasets}
\end{table}
\subsection{Evaluation measures}\label{subsec:eval_cri}
The performance is evaluated in terms of clustering by two widely used and standard measures, Accuracy (Acc) and Normalized Mutual Information (NMI). 
By taking $\mathbf{y}$ as the ground truth label information, and $\mathbf{z}$ as the predicted ones', Acc is defined as,
\begin{equation*}
Acc(\mathbf{y},\mathbf{z})=\frac{1}{n} \sum_{i=1}^{n} \delta(y_i,map(z_i)),
\end{equation*}
where $\delta(a,b)$ equals to 1 if $a=b$ and 0, otherwise. The best permutation of $\mathbf{z}$ to match $\mathbf{y}$ values is found by $map(.)$ function based on the Kuhn-Munkres approach \cite{lovasz_matching_1986}. The definition of NMI is given as,
\begin{equation*}
NMI(\mathbf{y},\mathbf{z}) = \frac{I(\mathbf{y},\mathbf{z})}{\max(H(\mathbf{y}),H(\mathbf{z}))},
\end{equation*}
where $H(.)$ represents the entropy and $I(\mathbf{y},\mathbf{z})$ is the mutual information of $\mathbf{y}$ and $\mathbf{z}$ defined as,
\begin{equation*}
I(\mathbf{y},\mathbf{z})=\sum _{y\in \mathbf{y}}\sum _{z\in \mathbf{z}}p(y,z)\log {\left({\frac {p(y,z)}{p(y)\,p(z)}}\right)}.
\end{equation*}

\subsection{The experimental setting}
The state-of-the-art UFS methods are applied such as LS\cite{he_laplacian_2005}, 
 UDFS\cite{yang_l21-norm_2011}, NDFS\cite{li_unsupervised_2012}, SPUFS \cite{lu_structure_2018}, LDSSL \cite{shang_local_2019}, and Baseline means to select all of the original features. 

We set $k=5$ on k-nearest neighbor algorithm, and $\sigma = 1$ for the bandwidth parameter in the Gaussian kernel for the methods based on explicit construction of the graph matrix. 
The $\gamma = 10^6$ is taken on our method and NDFS. The grid search strategy is employed to choose the appropriate weight parameters $\alpha$ and $\beta$ among the set of $\{10^{-4}, 10^{-2}, 1, 10^2, 10^4\}$ candidates. 
We limit data by selecting different number of features in the range of \{50, 100, 150, 200, 250, 300\} and cluster each ones by k-means algorithm, and then evaluate the clustering results by Acc and NMI measures. The mean and standard deviation values of Acc and NMI are reported by repeating the experiments for 20 times.
\subsection{Experimental results}
The performance of the feature selection algorithms are empirically evaluated in terms of Acc and NMI.
The mean and standard deviation of the clustering result are reported in the \mbox{Table \ref{tab:acc_res} and \ref{tab:nmi_res}}. The best and the second best results are marked as bold and underline. By considering the Table \mbox{\ref{tab:acc_res} and \ref{tab:nmi_res}}, we have the following conclusions,
\begin{itemize}
	\item The proposed approach outperform the Baseline method which is showed the efficacy of SCFS to select the more relevant features rather than the irrelevant and redundant ones'.
	\item Clustering based methods such as NDFS and SCFS commonly attain better results in an unsupervised manner.
\end{itemize}
%%%%%%%%AccuracyResultsTable
\begin{table}[h]
	\centering
	\caption{Clustering results (Acc\% $\pm$  std) of unsupervised feature selection methods on standard datasets. Bold and underlined numbers are the best and the second best.}
	\scriptsize
	\label{tab:acc_res}
	\begin{tabular}
		{
			l
			S[table-format=2.2]@{\,\(\pm\)\,}
			S[table-format=1.2]
			S[table-format=2.2]@{\,\(\pm\)\,}
			S[table-format=1.2]
			S[table-format=2.2]@{\,\(\pm\)\,}
			S[table-format=1.2]
			S[table-format=2.2]@{\,\(\pm\)\,}
			S[table-format=1.2]
			S[table-format=2.2]@{\,\(\pm\)\,}
			S[table-format=2.2]
			S[table-format=2.2]@{\,\(\pm\)\,}
			S[table-format=1.2]
		}
		\toprule
		Dataset & \multicolumn{2}{c}{Lung} & \multicolumn{2}{c}{Lymphoma} & \multicolumn{2}{c}{Prostate-GE} & \multicolumn{2}{c}{ORL}& \multicolumn{2}{c}{Isolet}& \multicolumn{2}{c}{BASEHOCK}  \\
		\midrule	
		Baseline  &71.67&6.86&58.75&5.19&58.82&0.00&\underline{59.14}&2.11&63.19&2.19&50.08&0.00\\
		LS			&61.46&2.61&50.12&2.26&60.82&1.71&49.94&3.62&53.31&4.49&50.63&0.23\\
		UDFS	 &56.65&4.93&59.60&2.50&\underline{61.36}&1.14&49.30&3.90&57.97&6.24&\underline{51.57}&0.56\\
		NDFS	 &\underline{83.71}&0.64&\underline{63.81}&0.33&60.07&0.77&58.81&1.19&\underline{67.72}&1.51&50.18&0.14\\
		SPUFS	&68.33&1.20&53.57&3.42&61.09&1.45&48.63&2.68&63.65&13.10&50.41&0.40\\
		LDSSL	&64.59&5.14&58.11&1.67&60.27&0.74&57.78&1.49&65.97&3.64&50.49&0.11\\
		SCFS	 &\textbf{86.70}&1.38&\textbf{64.87}&1.59&\textbf{61.70}&0.73&\textbf{59.19}&0.83&\textbf{69.17}&1.03&\textbf{51.95}&0.44\\
		\bottomrule
	\end{tabular}
\end{table}
%%%%%%%%%%%%%%%%%%%%%%%%%%%%%%%%%%%%%%%%%%%%%%%%
%%%%%%%%%%%%%%%%%%%%%%%%%%%%%%%%%%%%%%%%%%%%%%%%
\begin{table}[h!]
	\centering
	\caption{Clustering results (NMI\% $\pm$  std)  of unsupervised feature selection methods on standard datasets. Bold and underlined numbers are the best and the second best.}
	\scriptsize
	\label{tab:nmi_res}
	\begin{tabular}
		{
			l
			S[table-format=2.2]@{\,\(\pm\)\,}
			S[table-format=2.2]
			S[table-format=2.2]@{\,\(\pm\)\,}
			S[table-format=1.2]
			S[table-format=1.2]@{\,\(\pm\)\,}
			S[table-format=1.2]
			S[table-format=2.2]@{\,\(\pm\)\,}
			S[table-format=1.2]
			S[table-format=2.2]@{\,\(\pm\)\,}
			S[table-format=2.2]		
			S[table-format=1.2]@{\,\(\pm\)\,}
			S[table-format=1.2]
		}
		\toprule
		Dataset & \multicolumn{2}{c}{Lung} & \multicolumn{2}{c}{Lymphoma} & \multicolumn{2}{c}{Prostate-GE} & \multicolumn{2}{c}{ORL}& \multicolumn{2}{c}{Isolet}& \multicolumn{2}{c}{BASEHOCK}  \\
		\midrule
		Baseline	&62.90&2.76&68.95&3.63&2.55&0.00&\textbf{77.90}&0.86&77.61&1.12&0.63&0.00\\
		LS			  &50.16&5.95&55.88&2.52&4.46&1.65&70.90&2.67&70.41&4.52&\underline{2.54}&0.82\\
		UDFS	   &45.73&4.02&69.46&3.36&5.06&1.12&70.75&2.91&71.03&6.07&1.02&0.77\\
		NDFS	   &\underline{67.68}&0.85&\underline{73.70}&0.71&5.42&0.48&77.61&0.72&\underline{79.40}&1.72&1.20&0.81\\
		SPUFS	   &60.28&1.81&63.24&2.45&5.17&0.45&70.25&2.24&72.81&11.14&1.86&1.25\\
		LDSSL	  &52.31&5.19&64.64&1.66&\underline{5.66}&0.13&76.41&1.26&77.26&3.37&1.67&0.39\\
		SCFS		&\textbf{70.17}&0.90&\textbf{73.73}&0.75&\textbf{5.85}&0.46&\underline{77.71}&0.44&\textbf{79.43}&1.62&\textbf{3.73}&0.50\\
		\bottomrule
	\end{tabular}
\end{table}
%%%%%%%%%%%%%%%%%%%%%%%%%%%%%%%%%%%%%%%%%%%%%%%%

\begin{figure}[h!]
	\centering
	\begin{subfigure}[b]{.3\textwidth}
		\includegraphics[width=\textwidth]{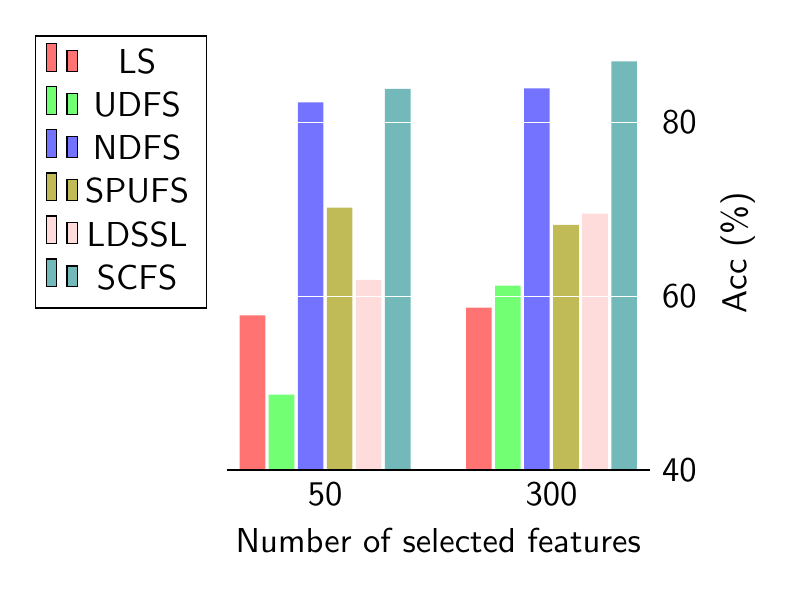}
		\caption{Lung}
		\label{fig:acc_Lung}
	\end{subfigure}
	\begin{subfigure}[b]{.3\textwidth}
		\includegraphics[width=\textwidth]{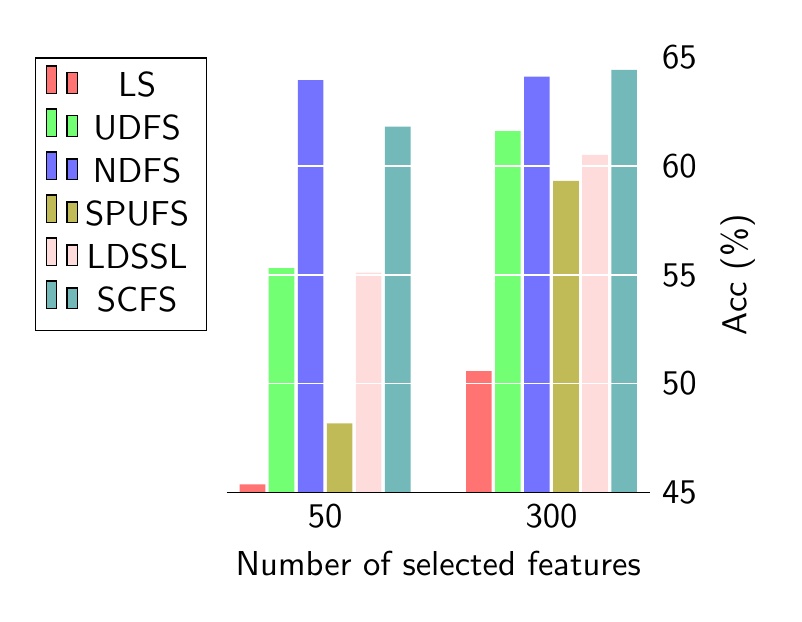}
		\caption{Lymphoma}
		\label{fig:acc_Lymphoma}
	\end{subfigure}
	\begin{subfigure}[b]{.3\textwidth}
		\includegraphics[width=\textwidth]{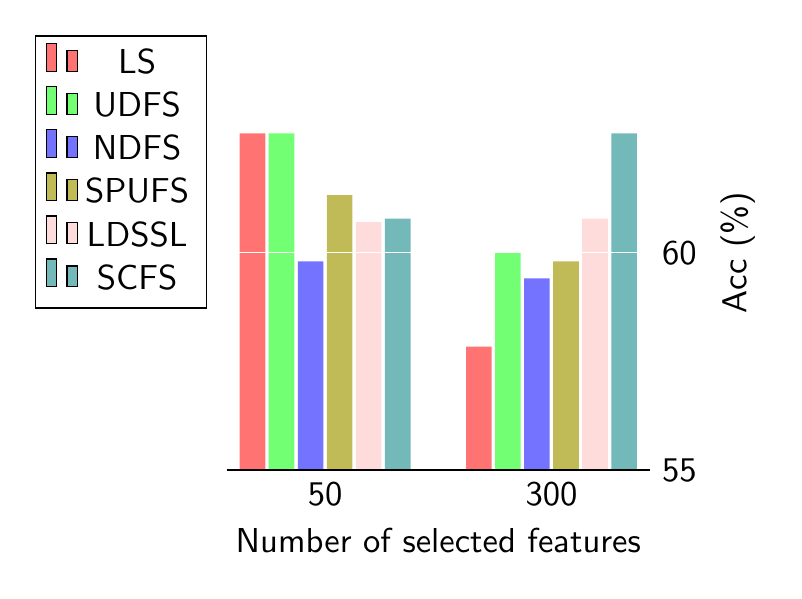}
		\caption{Prostate-GE}
		\label{fig:acc_Prostate}
	\end{subfigure}
	\begin{subfigure}[b]{.3\textwidth}
		\includegraphics[width=\textwidth]{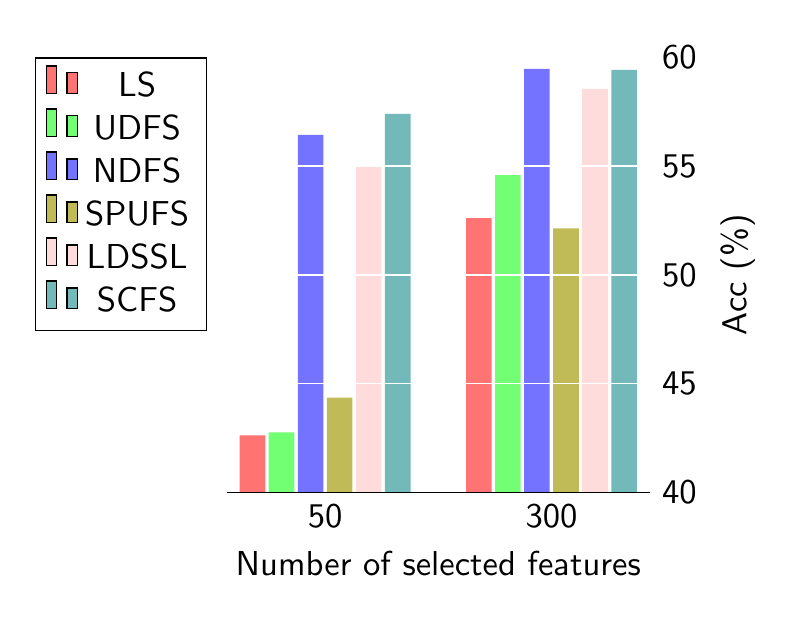}
		\caption{ORL}
		\label{fig:acc_ORL}
	\end{subfigure}
	\begin{subfigure}[b]{.3\textwidth}
		\includegraphics[width=\textwidth]{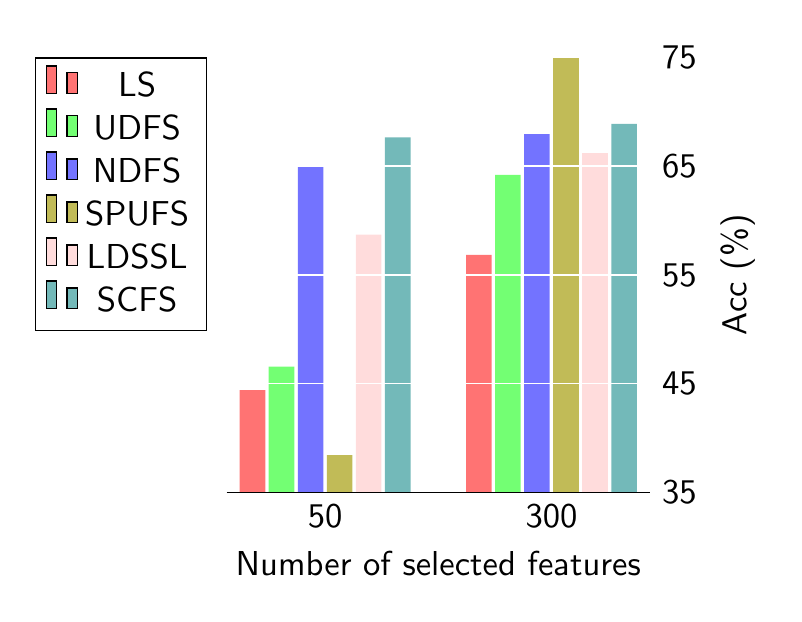}
		\caption{Isolet}
		\label{fig:acc_Isolet}
	\end{subfigure}
	\begin{subfigure}[b]{.3\textwidth}
		\includegraphics[width=\textwidth]{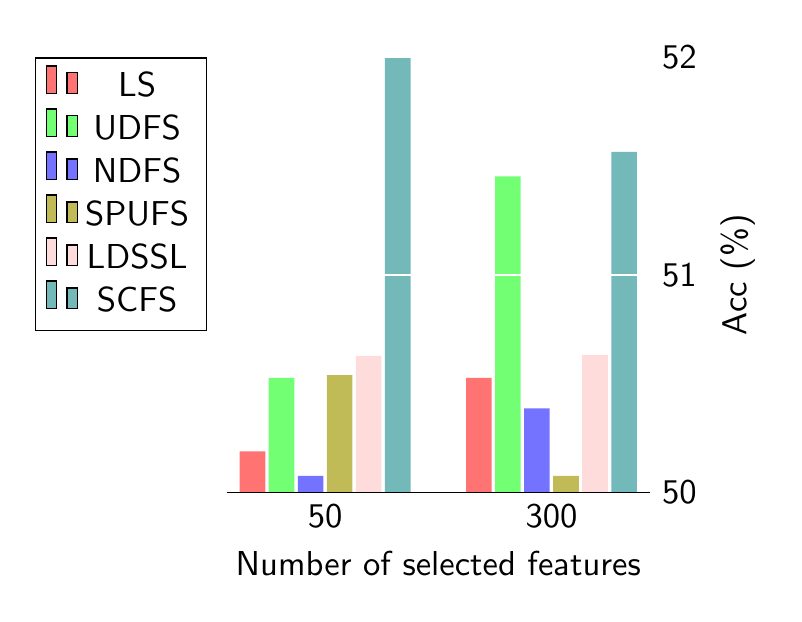}
		\caption{BASEHOCK}
		\label{fig:acc_BASEHOCK}
	\end{subfigure}
		\caption{The obtained results in terms of Acc with 50 and 300 numbers of selected features.}
		\label{fig:acc_barchart1}
\end{figure}

\begin{figure}[h!]
	\centering
	\begin{subfigure}[b]{.3\textwidth}
		\includegraphics[width=\textwidth]{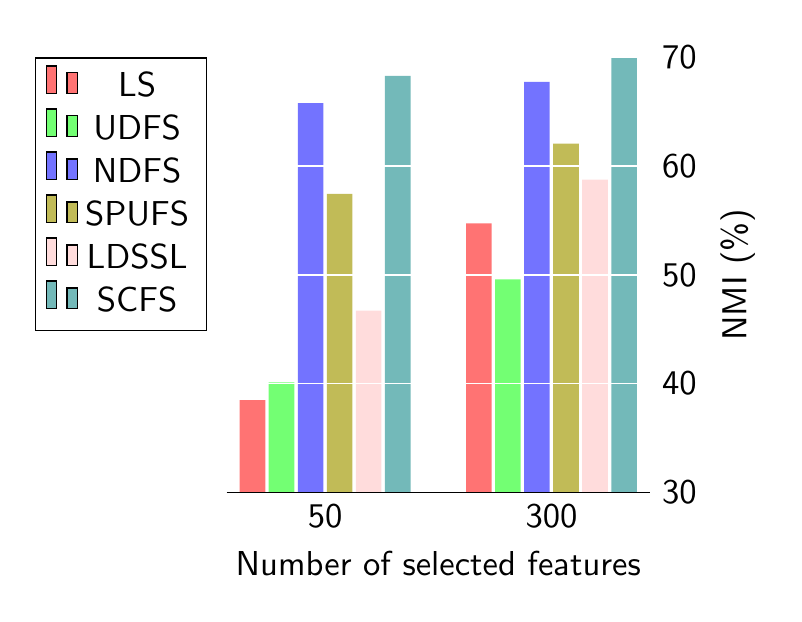}
		\caption{Lung}
		\label{fig:NMI_Lung}
	\end{subfigure}
	\begin{subfigure}[b]{.3\textwidth}
		\includegraphics[width=\textwidth]{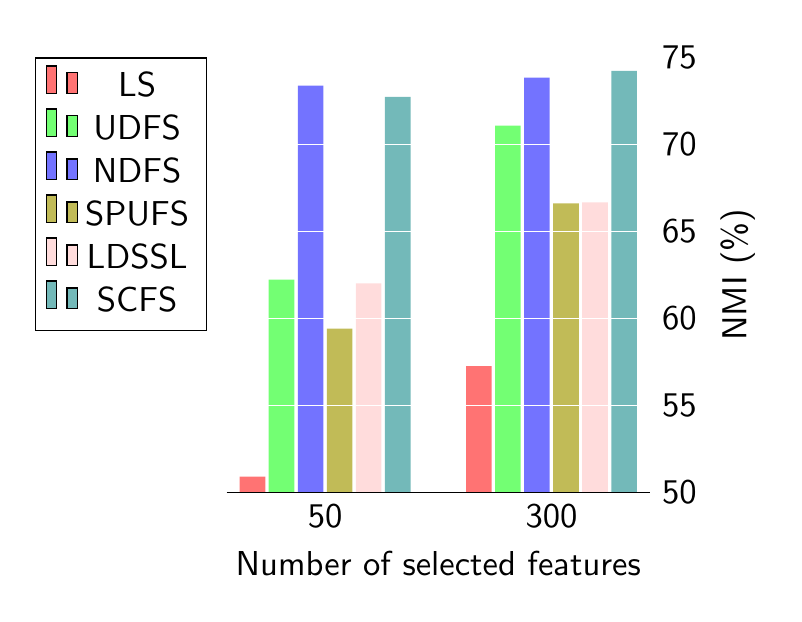}
		\caption{Lymphoma}
		\label{fig:NMI_Lymphoma}
	\end{subfigure}
	\begin{subfigure}[b]{.3\textwidth}
		\includegraphics[width=\textwidth]{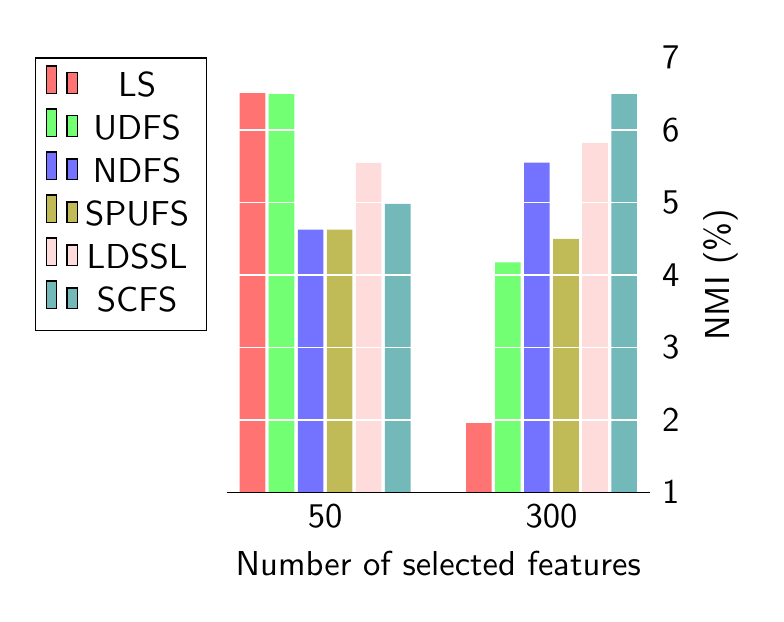}
		\caption{Prostate-GE}
		\label{fig:NMI_Prostate}
	\end{subfigure}
	\begin{subfigure}[b]{.3\textwidth}
		\includegraphics[width=\textwidth]{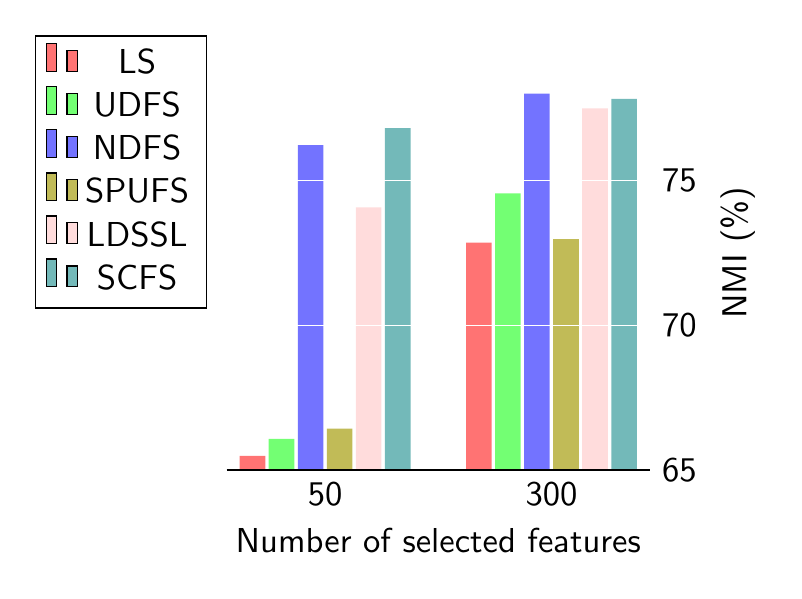}
		\caption{ORL}
		\label{fig:NMI_ORL}
	\end{subfigure}
	\begin{subfigure}[b]{.3\textwidth}
		\includegraphics[width=\textwidth]{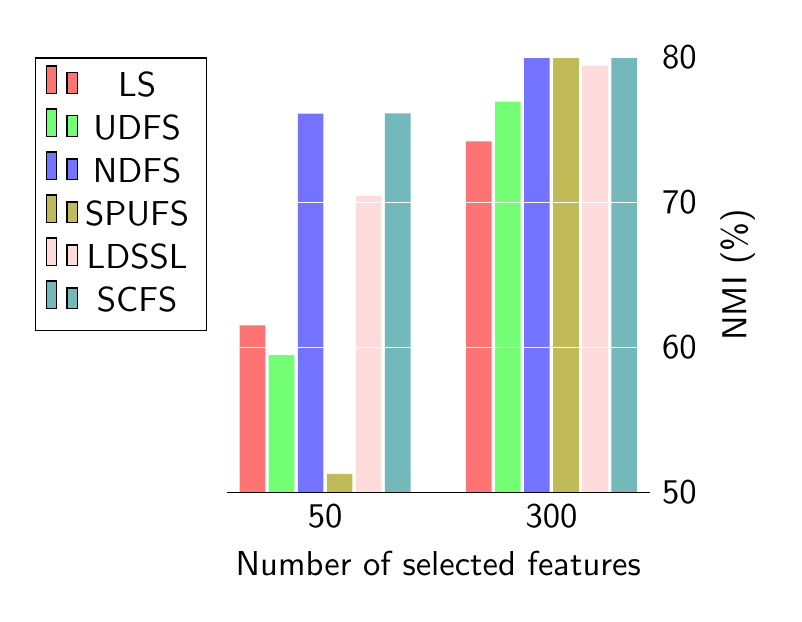}
		\caption{Isolet}
		\label{fig:NMI_Isolet}
	\end{subfigure}
	\begin{subfigure}[b]{.3\textwidth}
		\includegraphics[width=\textwidth]{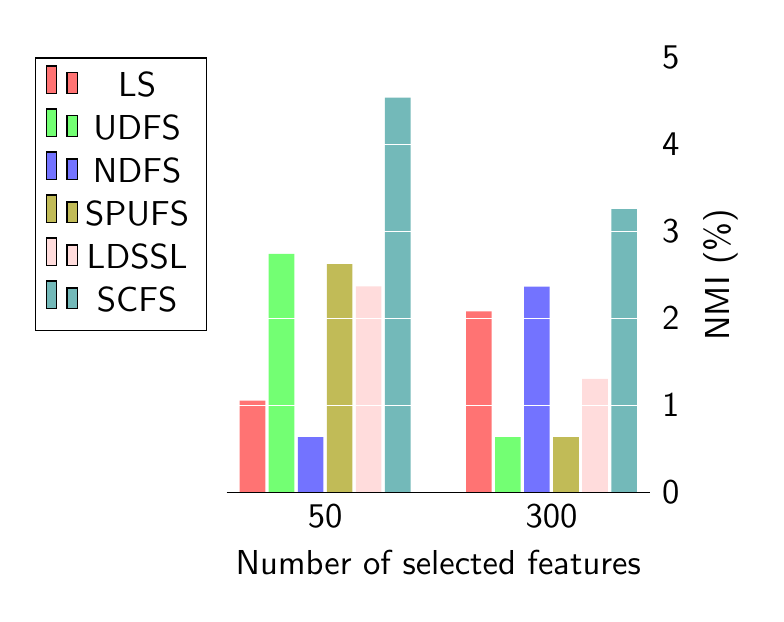}
		\caption{BASEHOCK}
		\label{fig:NMI_BASEHOCK}
	\end{subfigure}
	\caption{The obtained results in terms of NMI with 50 and 300 numbers of selected features.}
	\label{fig:NMI_barchart1}
\end{figure}

\begin{itemize}
	\item The proposed method, SCFS, achieves the best performance on Acc on the whole datasets, and also the best on NMI on the most cases.
	\item Moreover, the proposed method outperforms the earlier subspace learning based approach, LDSSL, due to employing the sample-level self-expression, and adaptive learning of the cluster similarities.  
\end{itemize}
Furthermore, we demonstrate the performance of the proposed method for two extreme scenarios, the first by considering the number of selected features as $50$, and the second as $300$. \mbox{Fig. \ref{fig:acc_barchart1} and Fig. \ref{fig:NMI_barchart1}} represent the obtained results according to these scenarios. 
On the one hand, SCFS performs satisfactory on the first scenario to deal with the small number of selected features.
On the other hand, the results indicate that the proposed approach attains better performance than the other well-known methods on almost all datasets on the second scenario.  
\clearpage

\subsection{Parameter sensitivity and convergence study}
First, the sensitivity of parameters $\alpha$ and $\beta$ in our model are investigated. The experimental results on Acc and NMI criteria for all of datasets are presented on \mbox{Fig. \ref{fig:sens_acc}}. 
For all candidate of $\alpha$ and $\beta$ parameters, the logarithms base 10 is taken. As shown in \mbox{Fig. \ref{fig:sens_acc}}, there is a relative sensitivity to the parameters, which is still an open problem. 

Next, we experimentally study the convergence behavior of the proposed algorithm.  \mbox{Fig. \ref{fig:convergence}} presents the speed of the convergence according to the objective values with respect to the number of iterations on different datasets. The stopping criteria is set as $\frac{obj(t)-obj(t-1)}{obj(t)}<10^{-5}$, where $obj(t)$ is the objective function value of \mbox{Eq. \eqref{eq:8}} in the $t$-th iteration. As shown in the \mbox{Fig. \ref{fig:convergence}}, the proposed algorithm monotonically decreases the objective function in a few iteration. 

\begin{figure}[t]
	\centering
	\begin{subfigure}[b]{0.32\textwidth}
		\includegraphics[width=\textwidth]{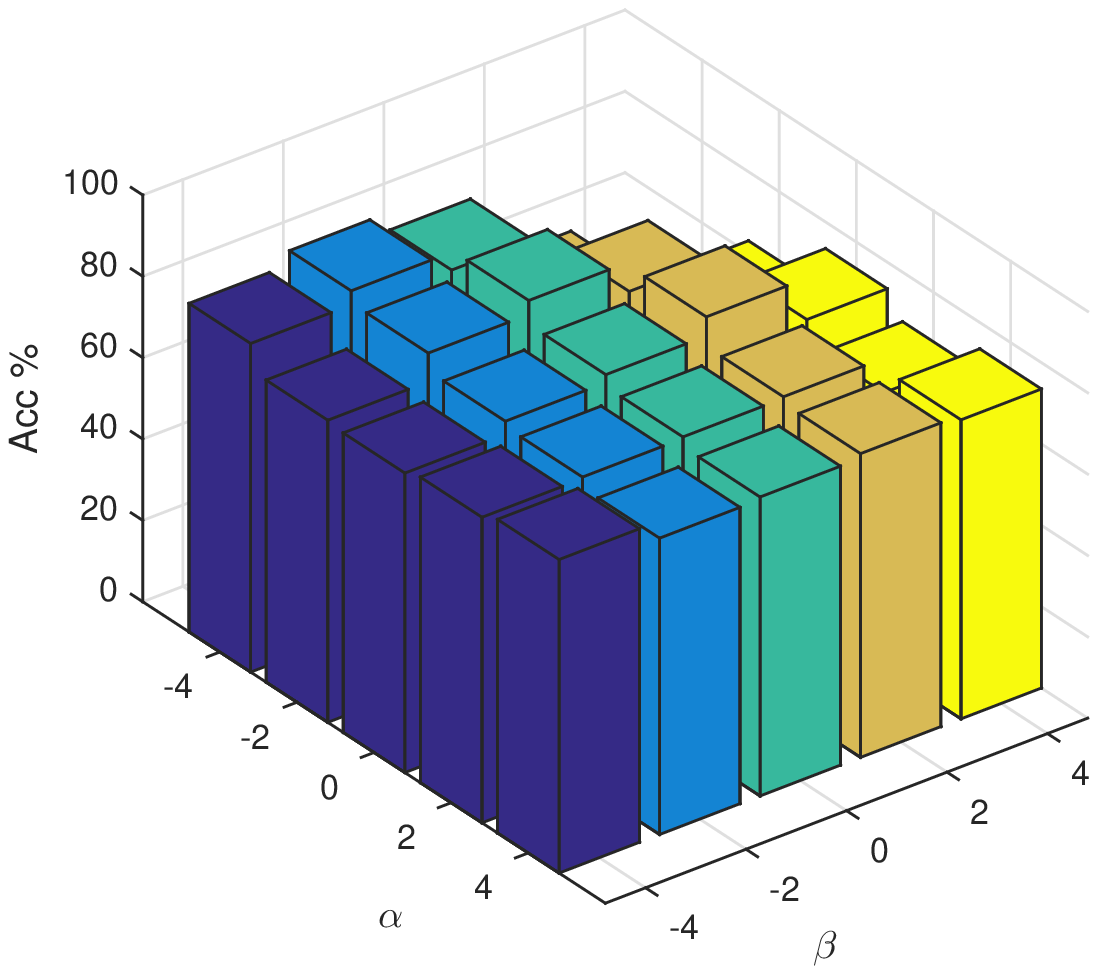}
		\caption{Lung}
		\label{fig:Lung_acc}
	\end{subfigure}
	\begin{subfigure}[b]{0.32\textwidth}
		\includegraphics[width=\textwidth]{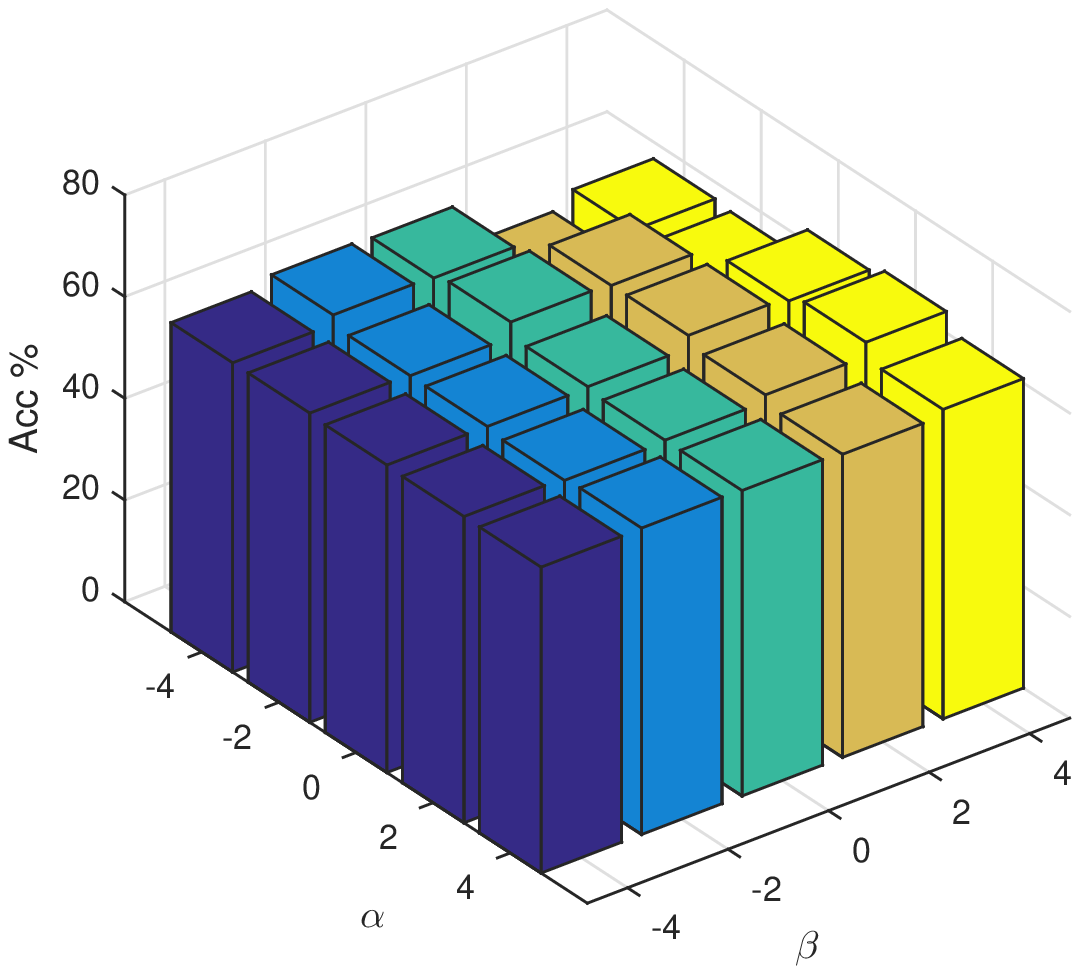}
		\caption{Lymphoma}
		\label{fig:Lymphoma_acc}
	\end{subfigure}
	\begin{subfigure}[b]{0.32\textwidth}
		\includegraphics[width=\textwidth]{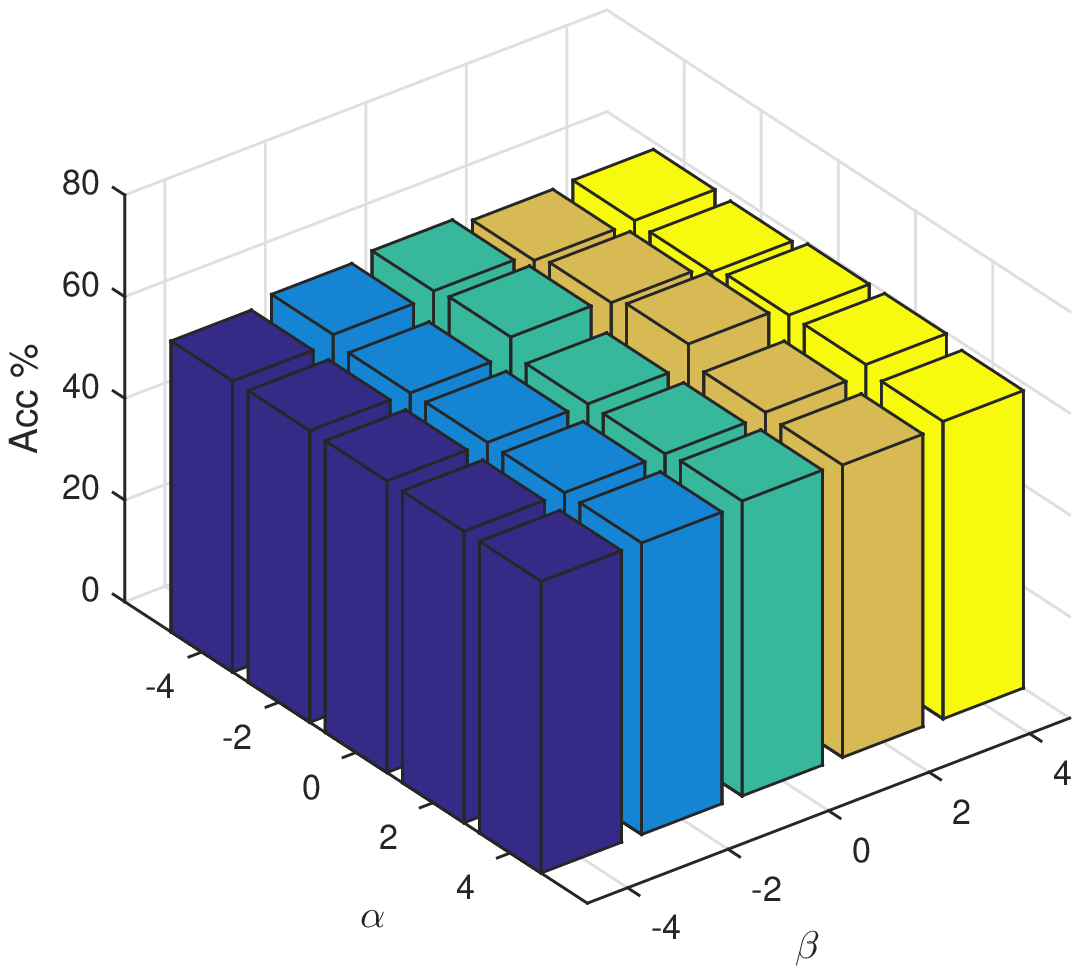}
		\caption{Prostate-GE}
		\label{fig:Prostate_acc}
	\end{subfigure}
	\begin{subfigure}[b]{0.32\textwidth}
		\includegraphics[width=\textwidth]{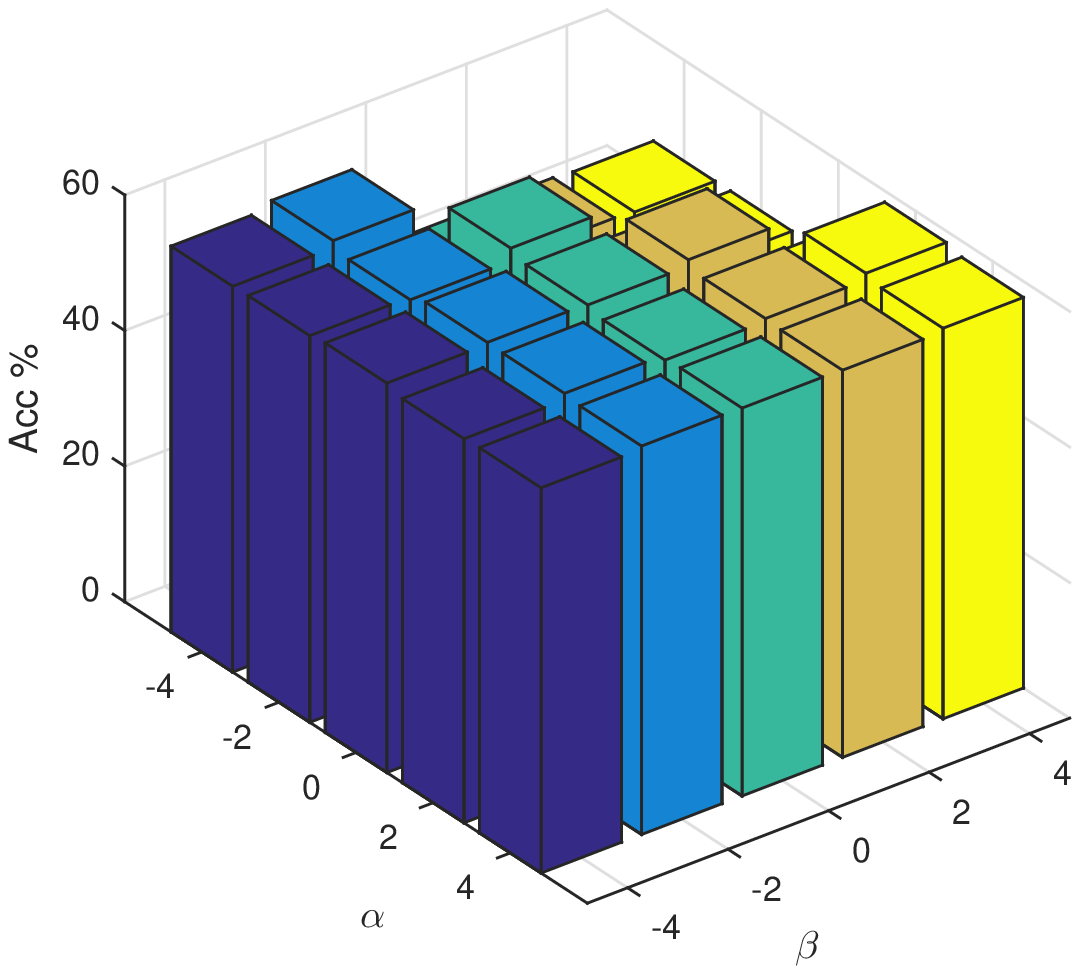}
		\caption{ORL}
		\label{fig:ORL_acc}
	\end{subfigure}
	\begin{subfigure}[b]{0.32\textwidth}
		\includegraphics[width=\textwidth]{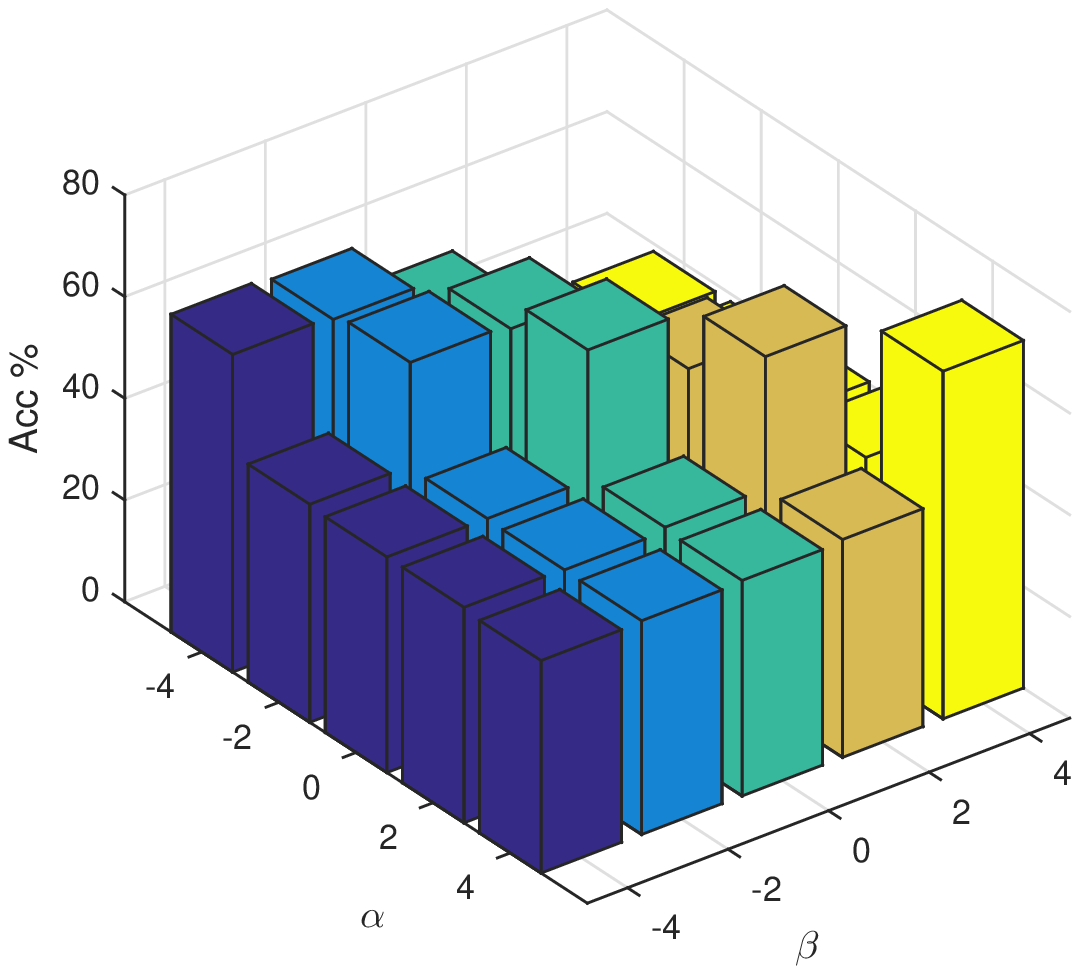}
		\caption{Isolet}
		\label{fig:Isolet_acc}
	\end{subfigure}
	\begin{subfigure}[b]{0.32\textwidth}
		\includegraphics[width=\textwidth]{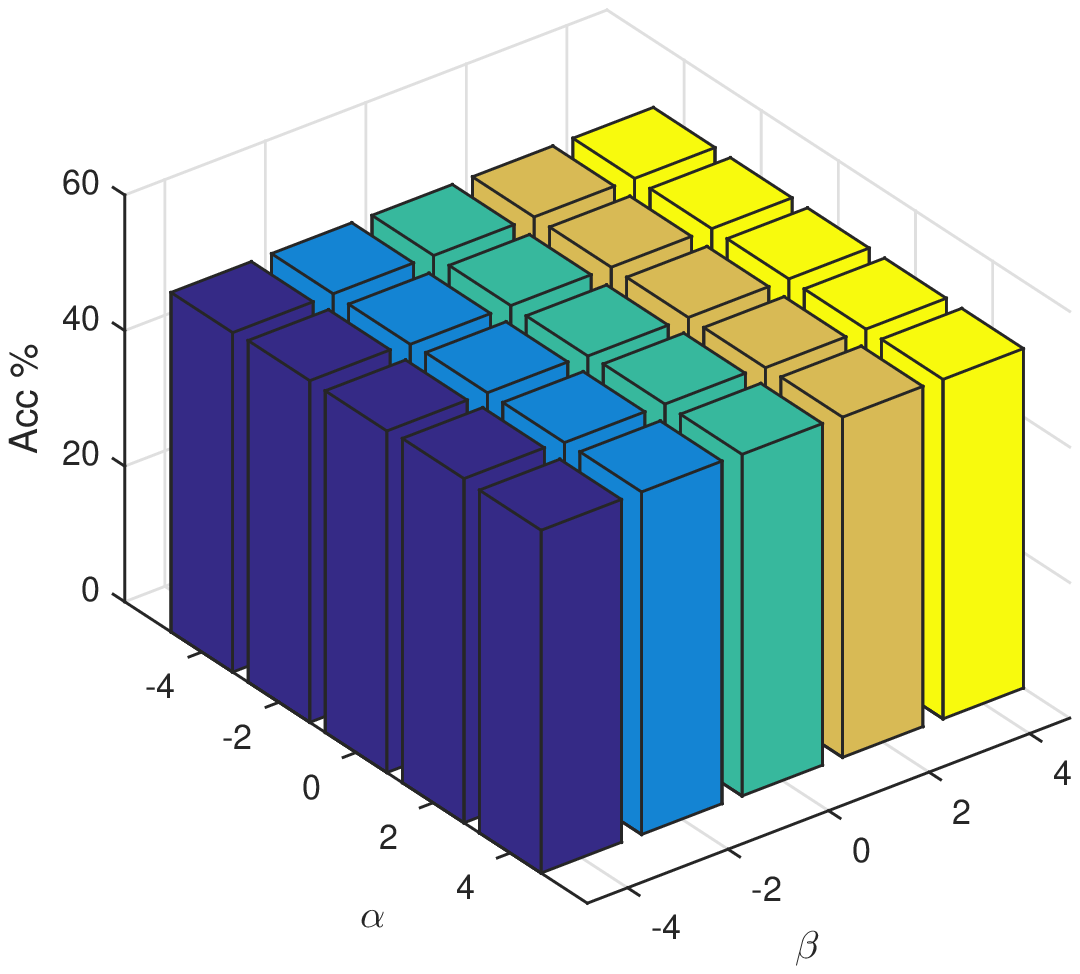}
		\caption{BASEHOCK}
		\label{fig:BASEHOCK_acc}
	\end{subfigure}
	\begin{subfigure}[b]{0.32\textwidth}
		\includegraphics[width=\textwidth]{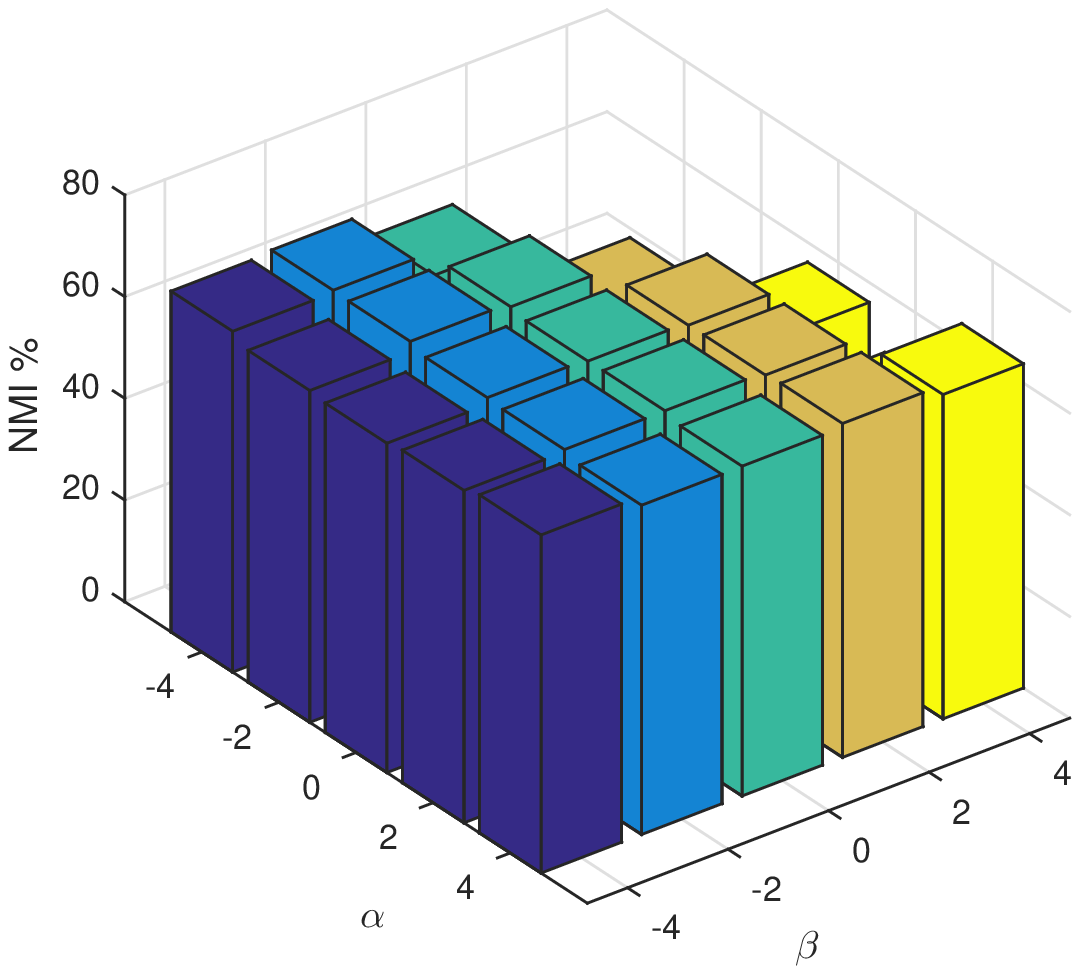}
		\caption{Lung}
		\label{fig:Lung_nmi}
	\end{subfigure}
	\begin{subfigure}[b]{0.32\textwidth}
		\includegraphics[width=\textwidth]{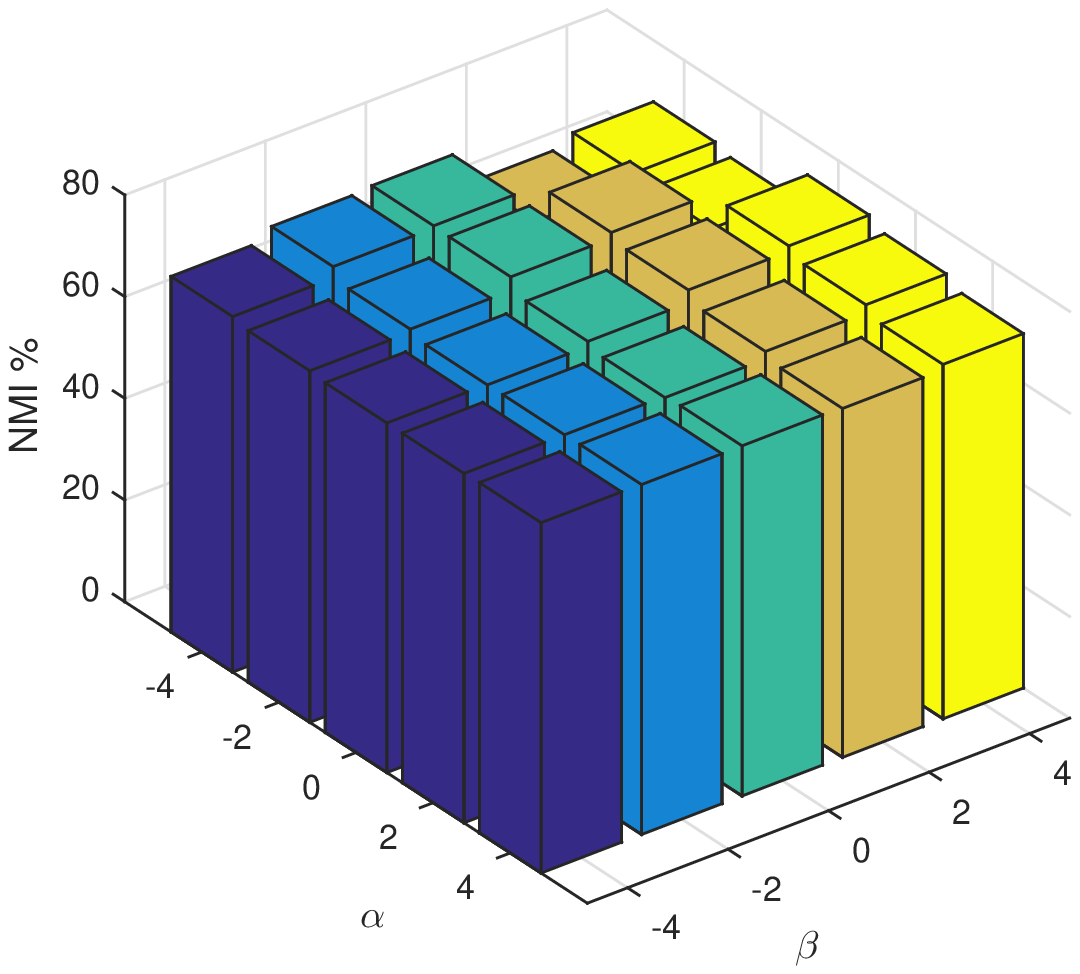}
		\caption{Lymphoma}
		\label{fig:Lymphoma_nmi}
	\end{subfigure}
	\begin{subfigure}[b]{0.32\textwidth}
		\includegraphics[width=\textwidth]{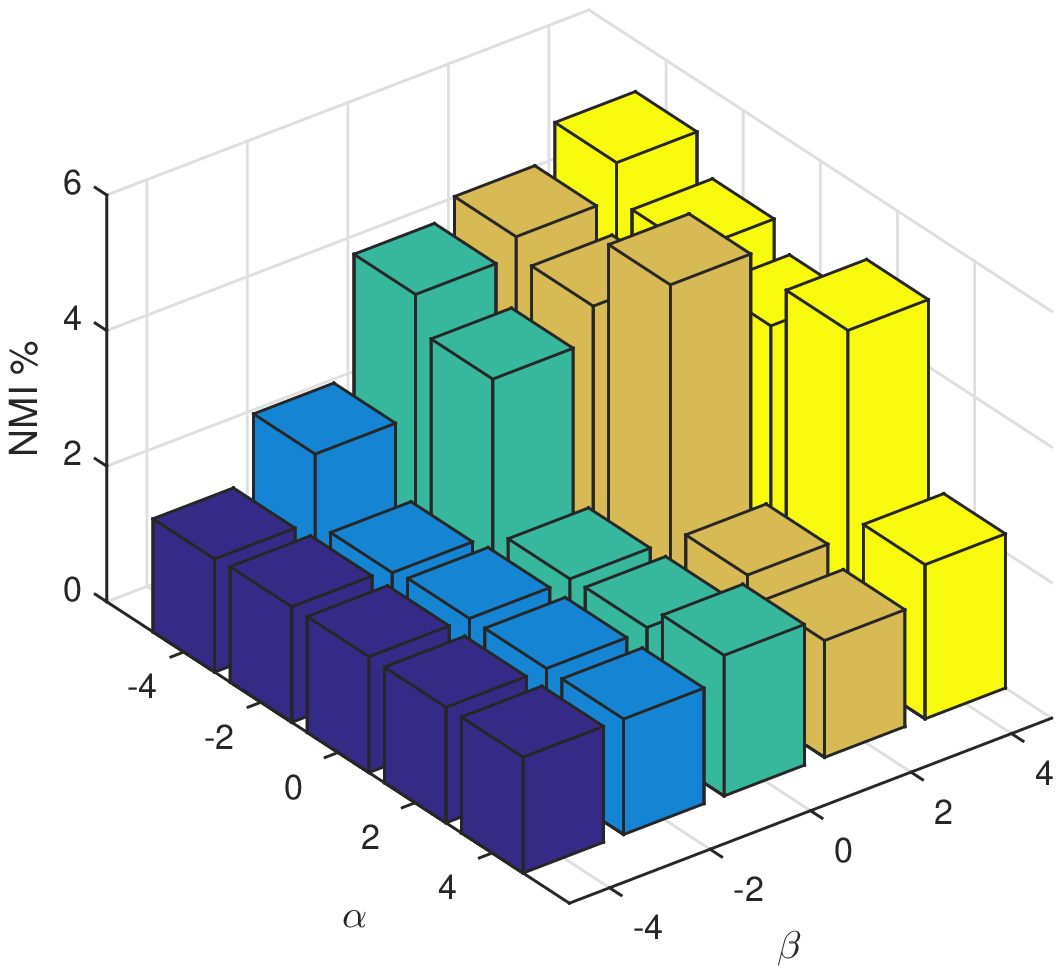}
		\caption{Prostate-GE}
		\label{fig:Prostate_nmi}
	\end{subfigure}
	\begin{subfigure}[b]{0.32\textwidth}
		\includegraphics[width=\textwidth]{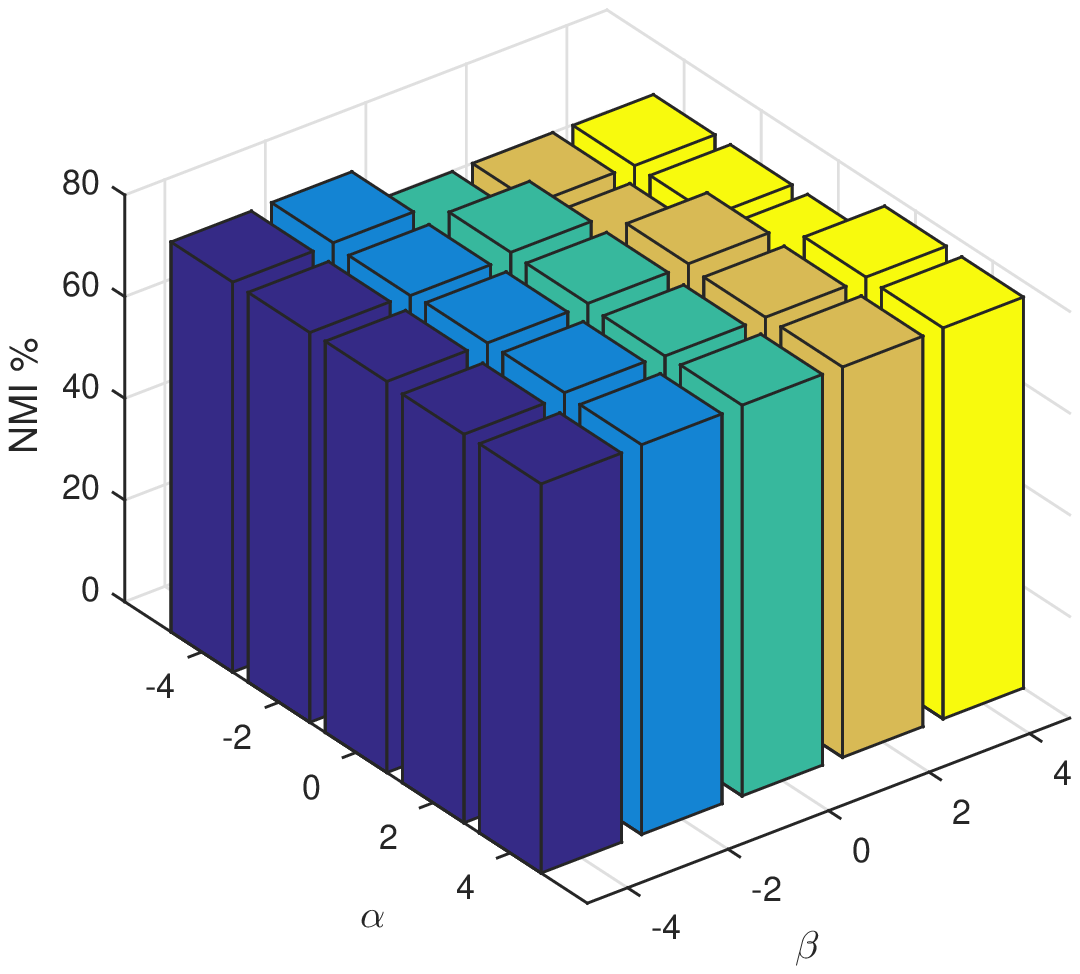}
		\caption{ORL}
		\label{fig:ORL_nmi}
	\end{subfigure}
	\begin{subfigure}[b]{0.32\textwidth}
		\includegraphics[width=\textwidth]{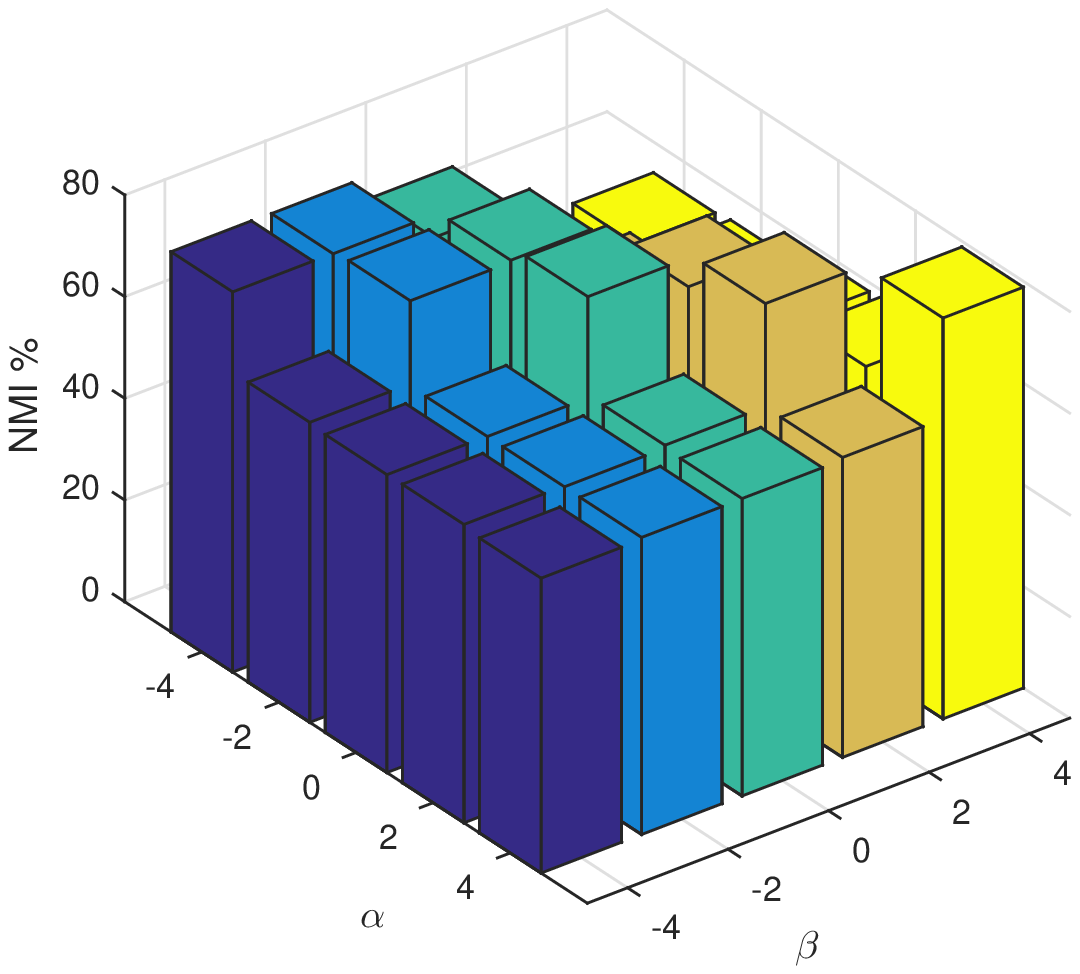}
		\caption{Isolet}
		\label{fig:Isolet_nmi}
	\end{subfigure}
	\begin{subfigure}[b]{0.32\textwidth}
		\includegraphics[width=\textwidth]{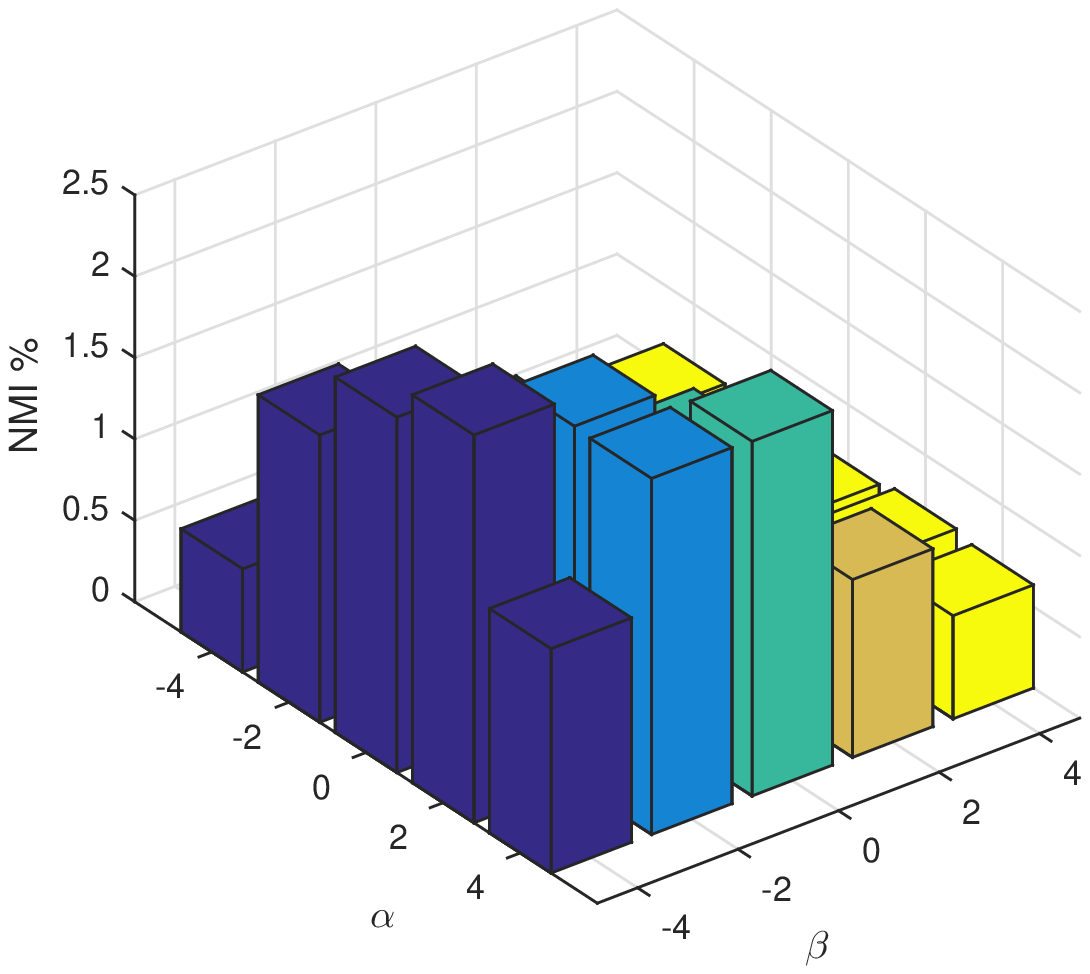}
		\caption{BASEHOCK}
		\label{fig:BASEHOCK_nmi}
	\end{subfigure}
	\caption{Acc and NMI of SCFS with different values of the parameters $\alpha$ and $\beta$.}
	\label{fig:sens_acc}
\end{figure}
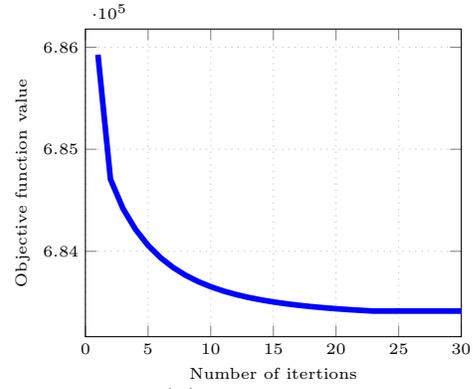
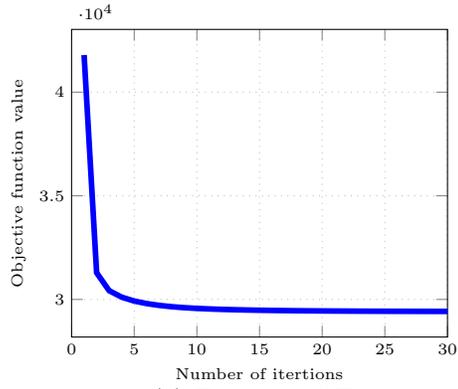
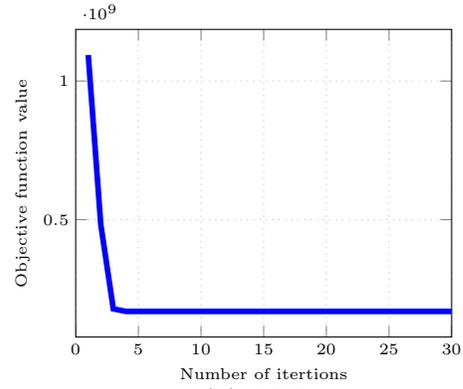
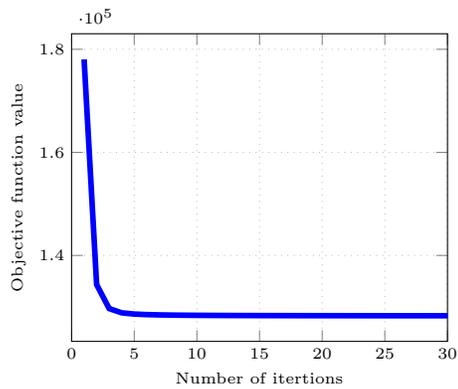
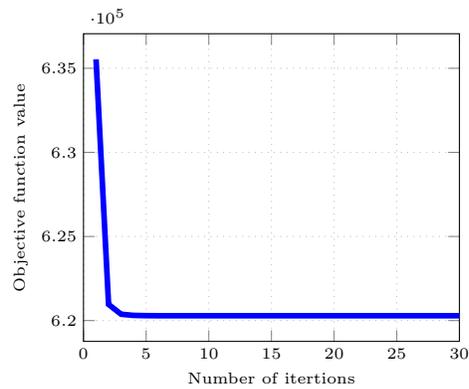
\begin{figure}[h!]
	\begin{subfigure}[b]{0.48\textwidth}
		\begin{tikzpicture}
		\tiny
		\begin{axis}[
		width=\linewidth, % Scale the plot to \linewidth
		grid=major, % Display a grid
		grid style={dotted}, % Set the style
		xlabel=Number of itertions, % Set the labels
		ylabel=Objective function value,
		xmin=0, xmax=30,
		xtick={0,5,10,15,20,25,30}
		]
		\addplot[no marks, color=blue, line width=0.75mm]
		coordinates {
			(1,60705.86284)(2,50483.25616)(3,49007.82947)(4,48387.82005)(5,48070.5968)(6,47897.38395)(7,47795.9844)(8,47732.52521)(9,47690.43747)(10,47661.17135)(11,47640.03801)(12,47624.30769)(13,47612.30617)(14,47602.96015)(15,47595.5549)(16,47589.59885)(17,47584.74462)(18,47580.7413)(19,47577.40431)(20,47574.59578)(21,47572.21135)(22,47570.17102)(23,47568.41278)(24,47566.88797)(25,47565.55802)(26,47564.39205)(27,47563.36508)(28,47562.45671)(29,47561.65014)(30,47560.93146)
		};
		\end{axis}
		\end{tikzpicture}
		\caption{Lung}
		\label{fig:Lung_con}
	\end{subfigure}\vspace{3mm}
	\begin{subfigure}[b]{0.48\textwidth}
		\begin{tikzpicture}
		\tiny
		\begin{axis}[
		width=\linewidth, % Scale the plot to \linewidth
		grid=major, % Display a grid
		grid style={dotted}, % Set the style
		xlabel=Number of itertions, % Set the labels
		ylabel=Objective function value,
		xmin=0, xmax=30,
		xtick={0,5,10,15,20,25,30}
		]
		\addplot[no marks, color=blue, line width=0.75mm]
		coordinates {
			(1,685927.002)(2,684706.6754)(3,684419.0371)(4,684214.8723)(5,684058.5095)(6,683936.9734)(7,683841.2288)(8,683764.8159)(9,683703.1127)(10,683652.772)(11,683611.333)(12,683576.9514)(13,683548.2211)(14,683524.0503)(15,683503.5819)(16,683486.1357)(17,683471.1704)(18,683458.2527)(19,683447.0346)(20,683437.2356)(21,683428.6281)(22,683421.0267)(23,683414.2795)(24,683414.2795)(25,683414.2795)(26,683414.2795)(27,683414.2795)(28,683414.2795)(29,683414.2795)(30,683414.2795)
		};
		\end{axis}
		\end{tikzpicture}
		\caption{Lymphoma}
		\label{fig:Lymphoma_con}
	\end{subfigure}\vspace{3mm}
	\begin{subfigure}[b]{0.48\textwidth}
		\begin{tikzpicture}
		\tiny
		\begin{axis}[
		width=\linewidth, % Scale the plot to \linewidth
		grid=major, % Display a grid
		grid style={dotted}, % Set the style
		xlabel=Number of itertions, % Set the labels
		ylabel=Objective function value,
		xmin=0, xmax=30,
		xtick={0,5,10,15,20,25,30}
		]
		\addplot[no marks, color=blue, line width=0.75mm]
		coordinates {
			(1,41790.63255)(2,31287.68228)(3,30423.3898)(4,30109.93854)(5,29923.48807)(6,29799.64247)(7,29712.5232)(8,29649.28711)(9,29602.71582)(10,29567.95982)(11,29541.45963)(12,29520.68951)(13,29503.98309)(14,29490.2953)(15,29478.96743)(16,29469.55237)(17,29461.71182)(18,29455.1681)(19,29449.68705)(20,29445.07272)(21,29441.16443)(22,29437.83274)(23,29434.9745)(24,29432.50776)(25,29430.36729)(26,29428.50079)(27,29426.866)(28,29425.42845)(29,29424.15976)(30,29423.03638)
		};
		\end{axis}
		\end{tikzpicture}
		\caption{Prostate-GE}
		\label{fig:Prostate_con}
	\end{subfigure}\vspace{3mm}
	\begin{subfigure}[b]{0.48\textwidth}
		\begin{tikzpicture}
		\tiny
		\begin{axis}[
		width=\linewidth, % Scale the plot to \linewidth
		grid=major, % Display a grid
		grid style={dotted}, % Set the style
		xlabel=Number of itertions, % Set the labels
		ylabel=Objective function value,
		xmin=0, xmax=30,
		xtick={0,5,10,15,20,25,30}
		]
		\addplot[no marks, color=blue, line width=0.75mm]
		coordinates {
			(1,1093680086)(2,482965143.9)(3,178339060)(4,169210519.9)(5,169202389.1)(6,169202389.1)(7,169202389.1)(8,169202389.1)(9,169202389.1)(10,169202389.1)(11,169202389.1)(12,169202389.1)(13,169202389.1)(14,169202389.1)(15,169202389.1)(16,169202389.1)(17,169202389.1)(18,169202389.1)(19,169202389.1)(20,169202389.1)(21,169202389.1)(22,169202389.1)(23,169202389.1)(24,169202389.1)(25,169202389.1)(26,169202389.1)(27,169202389.1)(28,169202389.1)(29,169202389.1)(30,169202389.1)
		};
		\end{axis}
		\end{tikzpicture}
		\caption{ORL}
		\label{fig:ORL_con}
	\end{subfigure}\vspace{3mm}
	\begin{subfigure}[b]{0.48\textwidth}
		\begin{tikzpicture}
		\tiny
		\begin{axis}[
		width=\linewidth, % Scale the plot to \linewidth
		grid=major, % Display a grid
		grid style={dotted}, % Set the style
		xlabel=Number of itertions, % Set the labels
		ylabel=Objective function value,
		xmin=0, xmax=30,
		xtick={0,5,10,15,20,25,30}
		]
		\addplot[no marks, color=blue, line width=0.75mm]
		coordinates {
			(1,178044.4004)(2,134347.1137)(3,129692.416)(4,128863.7527)(5,128623.8875)(6,128519.1701)(7,128460.4421)(8,128422.7184)(9,128396.5136)(10,128377.3638)(11,128362.8524)(12,128351.5454)(13,128342.5356)(14,128335.2218)(15,128329.1907)(16,128324.1497)(17,128319.8868)(18,128316.2447)(19,128313.1048)(20,128310.376)(21,128307.9875)(22,128305.8834)(23,128304.0192)(24,128302.3588)(25,128300.8729)(26,128299.5376)(27,128298.3327)(28,128298.3327)(29,128298.3327)(30,128298.3327)
		};
		\end{axis}
		\end{tikzpicture}
		\caption{Isolet}
		\label{fig:Isolet_con}
	\end{subfigure}\vspace{3mm}
	\begin{subfigure}[b]{0.48\textwidth}
		\begin{tikzpicture}
		\tiny
		\begin{axis}[
		width=\linewidth, % Scale the plot to \linewidth
		grid=major, % Display a grid
		grid style={dotted}, % Set the style
		xlabel=Number of itertions, % Set the labels
		ylabel=Objective function value,
		xmin=0, xmax=30,
		xtick={0,5,10,15,20,25,30}
		]
		\addplot[no marks, color=blue, line width=0.75mm]
		coordinates {
			(1,635530.3631)(2,620961.443)(3,620376.5216)(4,620307.4974)(5,620292.096)(6,620287.0765)(7,620287.0765)(8,620287.0765)(9,620287.0765)(10,620287.0765)(11,620287.0765)(12,620287.0765)(13,620287.0765)(14,620287.0765)(15,620287.0765)(16,620287.0765)(17,620287.0765)(18,620287.0765)(19,620287.0765)(20,620287.0765)(21,620287.0765)(22,620287.0765)(23,620287.0765)(24,620287.0765)(25,620287.0765)(26,620287.0765)(27,620287.0765)(28,620287.0765)(29,620287.0765)(30,620287.0765)
		};
		\end{axis}
		\end{tikzpicture}
		\caption{BASEHOCK}
		\label{fig:BASEHOCK_con}
	\end{subfigure}	
	\caption{Convergence curve of SCFS on different datasets.}
	\label{fig:convergence}
\end{figure}

\section{Conclusion}\label{sec:concl}
In this paper, we proposed a novel unsupervised feature selection framework initiated from the subspace learning and regularized regression to maintain sample similarities and take discriminative information into account in the selected features. The proposed method, SCFS, was designed to implicitly learn the cluster similarities in an adaptive manner.
Furthermore, a unified objective function was constituted from the main underlying characteristics of the proposed method. The optimization algorithm was proposed to obtain the solutions in an efficient way. In line with the computational complexity of the proposed algorithm, its convergence was investigated through an empirical study on real datasets. 
Extensive experiments on variaty of datasets was performed to show the effectiveness of the proposed method.
\section*{References}
\bibliographystyle{elsarticle-num} 
\bibliography{SCFS.bib}
\end{document}